# Spiking representation learning for associative memories

Naresh Ravichandran[1], Anders Lansner[1,2], Pawel Herman[1,3*]

[1]Computational Cognitive Brain Science Group, Department of Computational Science and Technology, School of Electrical Engineering and Computer Science, KTH Royal Institute of Technology, Stockholm, Sweden

[2]Department of Mathematics, Stockholm University, Stockholm, Sweden

[3]Digital Futures, KTH Royal Institute of Technology, Stockholm, Sweden

**\* Correspondence:**
Pawel Herman
paherman@kth.se

**Keywords: spiking neural networks, associative memory, attractor dynamics, Hebbian learning, structural plasticity, BCPNN, representation learning, unsupervised learning, machine learning, brain-like computing.**

## Abstract

Networks of interconnected neurons communicating through spiking signals offer the bedrock of neural computations. Our brain's spiking neural networks have the computational capacity to achieve complex pattern recognition and cognitive functions effortlessly. However, solving real-world problems with artificial spiking neural networks (SNNs) has proved to be difficult for a variety of reasons. Crucially, scaling SNNs to large networks and processing large-scale real-world datasets have been challenging, especially when compared to their non-spiking deep learning counterparts. The critical operation that is needed of SNNs is the ability to learn distributed representations from data and use these representations for perceptual, cognitive and memory operations. In this work, we introduce a novel SNN that performs unsupervised representation learning and associative memory operations leveraging Hebbian synaptic and activity-dependent structural plasticity coupled with neuron-units modelled as Poisson spike generators with sparse firing (~1 Hz mean and ~100 Hz maximum firing rate). Crucially, the architecture of our model derives from the neocortical columnar organization and combines feedforward projections for learning hidden representations and recurrent projections for forming associative memories. We evaluated the model on properties relevant for attractor-based associative memories such as pattern completion, perceptual rivalry, distortion resistance, and prototype extraction.

## 1 Introduction

The human brain has long captivated scientists and engineers across disciplines, serving as a wellspring of inspiration for advancements in artificial intelligence, robotics, computing paradigms, and algorithmic designs. The brain's remarkable efficiency, robustness, and parallel processing capabilities continues to act as a blueprint for developing sophisticated computing systems. Conventional computing paradigms, characterized by their sequential execution of instructions and rigid separation of memory and processing units, are increasingly being challenged by the growing demands of emerging applications such as real-time data analytics, autonomous systems, and cognitive computing. The brain seamlessly integrates memory and computation, operates in a massively parallel fashion, and exhibits remarkable fault tolerance and energy efficiency – all features that modern computing systems strive to achieve.

The potential of brain-like computing is evident in the realm of SNNs with the aim to reduce the energy cost. Notably, this energy efficiency of the brain is not attributed to a small network size; rather, the human brain packs in billions of neurons and trillions of synapses. SNNs have been shown to efficiently process real-time data streams through sparse and asynchronous event-based communication paradigms (Roy et al., 2019; Marković et al., 2020; Zenke and Neftci, 2021; Schuman et al., 2022). However, despite their potential, SNNs currently face several limitations. Notably, they lack robust mechanisms for learning sparse distributed internal representations from real-world data, a capability essential for real-world pattern recognition tasks, as deep learning has demonstrated. Moreover, in the spirit of human-like perceptual functionality, SNNs should be able to learn these representations in an unsupervised manner and utilize it for associative memory function, a hallmark feature of neural computations in the brain. Addressing these challenges is crucial for unlocking the full potential of SNNs and their neuromorphic implementations.

In this work, with the ambition to systematically tackle the aforementioned challenges, we introduce and evaluate a novel SNN model grounded in our previous work on non-spiking brain-like computing architectures (Ravichandran et al., 2020, 2021, 2023a, 2023b, 2024). Our earlier work derived from the Bayesian Confidence Propagation Neural Network (BCPNN) framework (Lansner and Ekeberg, 1989) and showed capacity to learn sparse distributed representations and employ these representations for associative memory function. Here, our spiking neuron model is a stochastic Poisson spike generation process and operates at low firing rates recapitulating the characteristics of *in vivo* cortical pyramidal neurons. We have incorporated several brain-like features into our model to enhance its biological plausibility (O'Reilly, 1998; Pulvermüller et al., 2021; Ravichandran et al., 2024): (1) Hebbian plasticity: online synaptic learning leveraging only localized correlational information from pre- and post-synaptic spikes, (2) Structural plasticity: an activity-dependent rewiring algorithm that learns a sparse (<10%) patchy connectivity matrix, (3) Sparsely spiking activities: neuronal firing with Poisson statistics and around 1 Hz mean and 100 Hz maximum firing rate, (4) Neocortical columnar architecture: functional hypercolumn modules with minicolumns competing in a soft-winner-takes-all manner, and (5) Cortex-like network architecture: feedforward, recurrent, and feedback projections. Crucially, our feedforward projections are responsible for extracting sparse distributed hidden representations from data and the recurrent projections facilitate robust and reconstructive associative memory functions through attractor dynamics.

Based on the results from our previous research (Ravichandran et al., 2020, 2021, 2023a, 2024), here in this work we have tested the following hypothesis: *the sparsely spiking Poissonian neurons integrated within our brain-like network architecture achieve the same performance as non-spiking rate-based networks in terms of learning representations and associative memory functionality*. To this effect, we designed models with feedforward-only and full architectures (*Ff* and *Full*; Fig. 1a), each with three different activations (Fig. 1b): rate-based (*Rate*), spiking (*Spk*; 1000 Hz maximum firing rate)*,* and sparsely spiking (*Spspk;* 100 Hz maximum firing rate). In effect, we have compared the following six models:

1. *RateFf:* rate-based activation in a feedforward network (without recurrent projections)
2. *RateFull*: rate-based activation in a full network (with recurrent projections)
3. *SpkFf*: spiking activation in a feedforward network
4. *SpkFull*: spiking activation in a full network
5. *SpspkFf*: sparsely spiking activation in a feedforward network, and
6. *SpspkFull*: sparsely spiking activation in a full network.

We have evaluated our models on the widely used MNIST hand-written digits dataset and made the following key observations: (1) The sparsely spiking model closely approximates the spiking (densely spiking) and rate-based models in terms of representation learning and associative memory function; (2) The previous published rate-based BCPNN model can be recast entirely as a sparsely spiking model with minimal modifications, and a synaptic short-term filtering (*z*-traces) is sufficient and necessary for this procedure; (3) The addition of recurrent projections enable the model to perform associative memory function and render it more robust compared to a feedforward-only model.

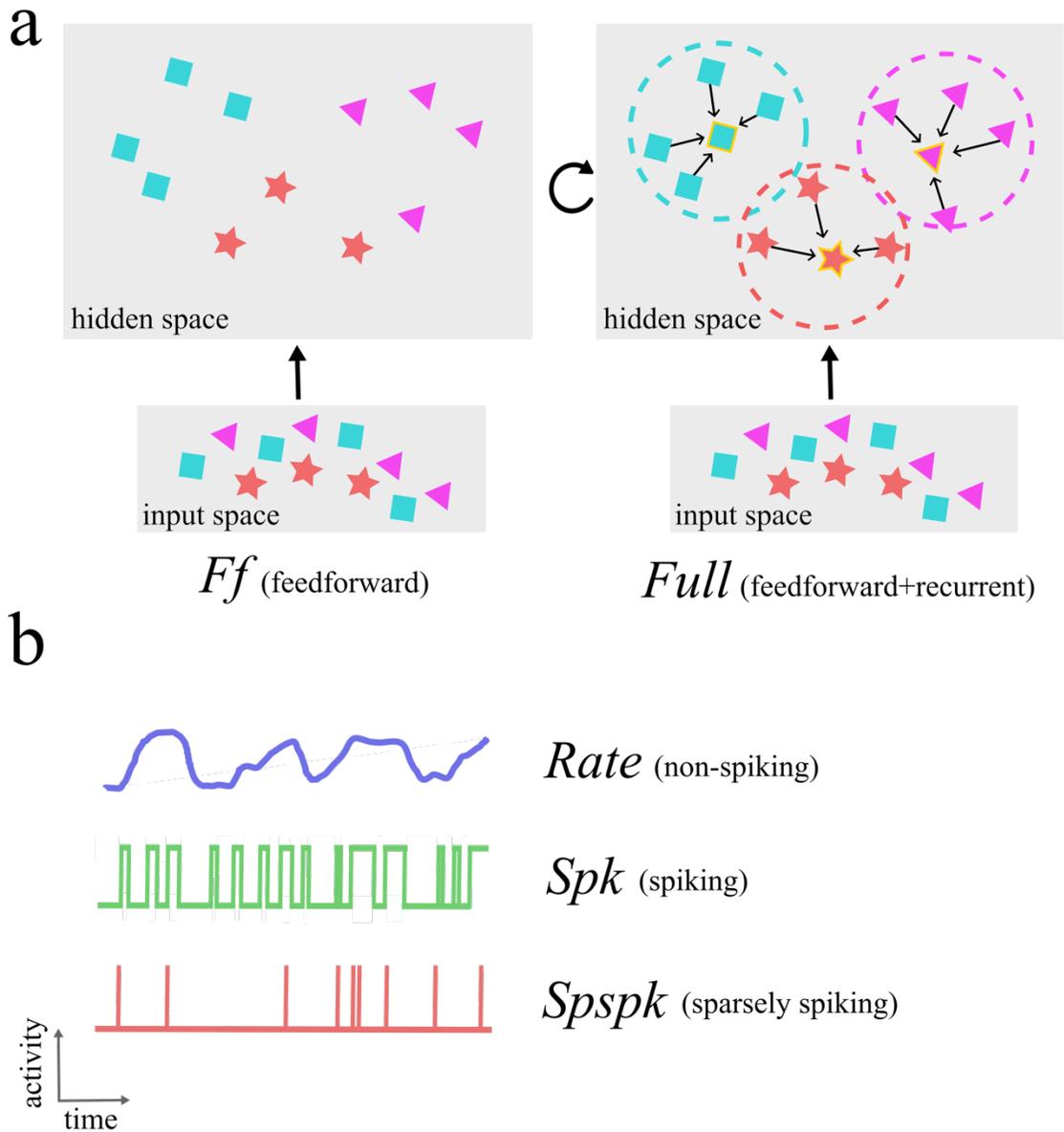

*Figure 1. Conceptual schematic of functional roles of the different architectures and neuron-unit activation types investigated in this work. (a) In the feedforward-only network (Ff), the representations in the input space are highly correlated and data from distinct categories and with different features are entangled in complex non-linear relationships (shown as purple triangles, red star, and blue squares). The feedforward projections learn to map these data into the hidden space where the data points are less correlated and grouped together based on the feature similarity making them more linearly separable. The Full architecture that includes the recurrent projection utilizes the uncorrelated nature of the representations in the hidden space to form effective associative memories and group similar data points into attractors (attractor boundaries or basins of attraction shown as dashed circles and attractor states as symbol with golden border). (b) The activation function denoting the signal computed and communicated by each neuron-unit can be one of either Rate (non-spiking), Spk (spiking), or Spspk (sparsely spiking). The rate-based activation codes for the probability of the presence of a feature that the neuron-unit represents ("confidence") and takes continuous values in*

*the interval [0,1]. The Spk activation is generated as stochastic binary samples from the underlying firing rate which can reach biologically implausible levels up to 1000 Hz. The Spspk activation is generated as stochastic binary samples, but with firing rate scaled down to biologically realistic values with the maximum of 100 Hz.*

## 2  Related works

### 2.1  Models of associative memory and their limitation when learning correlated memories

The synaptic connections in the brain, especially in the neocortex, are found to be predominantly recurrent in nature (Douglas and Martin, 2007). Yet their precise role in cortical information processing remains unclear (Spoerer et al., 2017; Kar et al., 2019; Kietzmann et al., 2019a; van Bergen and Kriegeskorte, 2020). One prominent hypothesis suggests that extensive recurrence facilitates associative memory, wherein distributed assemblies of coactive neurons reinforce each other (Willshaw et al., 1969; Palm, 1980; Lansner, 2009). This concept of cell assembly, known variously as associative memory (Hebb, 1949; Lansner et al., 2003; Harris, 2005), attractor (Hopfield, 1982; Amit, 1989; Khona and Fiete, 2022), ensemble (Yuste et al., 2024), avalanche (Plenz and Thiagarajan, 2007), cognit (Fuster, 2006) among others, is hypothesized to serve as the internal representations of memorized objects. Several theoretical and computational studies have shown that recurrently connected neuron-like binary units with symmetric connectivity can implement attractor dynamics: the network is guaranteed to converge to attractor states corresponding to local energy minima in analogy with statistical physics (Hopfield, 1982; Amit, 1989). Learning memories in such networks typically follows Hebbian synaptic plasticity, i.e., the synaptic connections between neurons are strengthened when they are coactive (and weakened otherwise). Subsequent work showed recurrent modular networks where each module can be in one of many possible discrete states have increased storage capacity compared to non-modular networks (Kanter, 1988; Gripon and Berrou, 2011; Knoblauch and Palm, 2020).

Associative memories functionally reflect the Gestalt nature of perception of the whole form rather than just a collection of isolated parts (Wagemans et al., 2012) and the reconstructive nature of memory discussed in psychology (Anderson and Bower, 1973; Bartlett and Kintsch, 1995). We describe four key functions of associative memories (Palm, 1980; Lansner, 2009; Rolls and Treves, 2012):

1. *Prototype extraction* relates to psychological studies on concept formation and categorical knowledge representation where concepts are stored as a set of descriptors of a prototype and novel examples are judged to be category members based on their closeness to this prototype (Rosch, 1988). The concept representations engage large-scale patterns of neural activity distributed across the neocortex and hippocampus (Kiefer and Pulvermüller, 2012; Handjaras et al., 2016; Fernandino et al., 2022).
2. *Pattern completion* involves reconstructing a complete memory pattern when cued with partial patterns. As a memory related phenomenon, it is particularly associated with the hippocampus and neocortex in humans (Horner et al., 2015; Liu et al., 2016). Interestingly, in human visual perception tasks with partially occluded objects, behavioral choices as well as neuronal dynamics show delayed responses (ca. 50-100 ms) compared to when whole objects are presented, which suggests the involvement of recurrent and top-down processing (Tang et al., 2014, 2018). Mice studies have shown that optogenetic stimulation of a specific subset of neurons belonging to an ensemble gives rise to the recall of the whole pattern (Carrillo-Reid et al., 2016) and causally trigger behavioral responses even in the absence of visual stimulation (Carrillo-Reid et al., 2019).

3. *Pattern rivalry* corresponds to scenarios where multiple conflicting pattern cues (typically two) are simultaneously presented and only one of these competing patterns "wins over" as a percept. Pattern rivalry phenomena is studied in the psychological domain such as the Necker cube and face-vase illusions where multiple distinct object representations compete for perceptual awareness (Carter et al., 2020). In binocular rivalry, two dissimilar images simultaneously presented to each eye compete for perceptual awareness (Lumer et al., 1998; Blake and Logothetis, 2002).

4. *Distortion resistance* implies the capability of the network to reconstruct the original pattern even if presented with a distorted cue, e.g., due to noise, limited viewing angle, poor contrast or illumination (Ghodrati et al., 2014; Wichmann et al., 2017; Geirhos et al., 2018a, 2018b).

While associative memory networks have been successful in modeling cortical dynamics and memory phenomena, they have primarily been trained on artificially generated orthogonal or random patterns (Hopfield, 1982; Rolls and Treves, 2012). Attractor networks struggle to reliably store overlapping (non-orthogonal) patterns, typical of real-world datasets, as they cause memory interference, so-called crosstalk, and lead to the emergence of spurious memories (Amit et al., 1987). This is a severe issue for considering associative memory networks as models of brain computation since the brain deals with high-dimensional sensory input with complex correlations. Consequently, attractor networks have not really been combined with high-dimensional correlated input and the problem of extracting suitable representations from real-world data has not received much attention in the context of associative memory. The associative memory systems in the brain (higher-order cortical associative areas and hippocampus, for instance) evidently use highly transformed representations. It is hypothesized that desirable neural representations in the brain are extracted from sensory input by feedforward cortical pathways (Felleman and Van Essen, 1991; Fuster, 2006; DiCarlo et al., 2012).

## 2.2 Representation learning algorithms and issues in transferring them to the spiking domain

The question of the nature of representations to be extracted from data has been studied under the topic of representation learning in the brain and in computational models (Bengio et al., 2013). Biological inspiration has been loosely adopted in deep neural networks (DNN) developed for pattern recognition on complex datasets, e.g., natural images, videos, audio, natural languages (LeCun et al., 2015). The success of deep learning in solving various real-world pattern recognition benchmarks has showcased the importance of learning distributed internal representations.

Compared to deep learning models SNNs still lack in their representation learning capacity. Building such SNNs has been typically addressed either by converting a (non-spiking) deep neural network model trained with gradient descent into a SNN, or by modifying supervised backprop-based gradient descent algorithms to accommodate spiking neurons (Roy et al., 2019; Wunderlich and Pehle, 2021; Zenke and Neftci, 2021; Cramer et al., 2022; Eshraghian et al., 2023). This approach has the advantage of exploiting the powerful gradient-based optimization techniques that have been developed extensively for DNNs. However, it is not straightforward to convert gradient-based backprop learning to a spiking domain, since spiking activation does not comply well with continuous differentiable activation function that backprop builds on. Several recent works have shown how the spiking activation can be smoothened into an activation function suitable for backprop and these models have demonstrated considerable success (Cramer et al., 2022). However, these methods typically carry many of the limitations of deep learning such as being predominantly supervised in their training, long training iterations, sensitivity to out-of-training noise, etc. Another critical issue with this approach is that it does not shed light on the learning in the brain and loses out on the impressive qualities that accompany a brain-like approach.

One prominent brain-like approach to learn hidden representations is to use biologically plausible spiking neuron activations and a localized form of learning rules. Early studies showed individual non-spiking neurons can develop selectivity to specific features when the hidden layer employs winner-takes-all competition (Rumelhart and Zipser, 1985; Linsker, 1988; Sanger, 1989; Bell and Sejnowski, 1995; Rozell et al., 2008). Later work incorporated spiking neurons and spike-timing-dependent plasticity (STDP) for learning and applied it to image recognition benchmarks (Masquelier and Thorpe, 2007; Diehl and Cook, 2015). The aforementioned models were restricted to hidden layers with a global winner-takes-all competition which makes each neuron learn exclusive features from the data, typically prototype clusters, and form localist coding. However, for a fully distributed spiking representation where neurons code for non-exclusive local features from the data, the hidden layer constituting multiple modules each with winner-takes-all competition was shown to learn distributed representations and perform well on machine learning benchmarks (Roy and Basu, 2017; Pfeiffer and Pfeil, 2018; Tavanaei et al., 2019; Taherkhani et al., 2020; Ravichandran et al., 2023c).

## 2.3 Complex network architectures integrating feedforward and recurrent projections

Neural network models can have complex architectures that combine the capacity of feedforward models to learn sparse distributed representations and employ them in a recurrent setting to form associative memory functions. Such architectures have been explored recently and benchmarked on machine learning datasets (Wyatte et al., 2012; O'Reilly et al., 2013; Tang et al., 2018, 2023; Kar et al., 2019; Kietzmann et al., 2019b; Sa-Couto and Wichert, 2020; Salvatori et al., 2021, 2024; Ravichandran et al., 2023a; Sacouto and Wichert, 2023; Simas et al., 2023a). O'Reilly et al., (2013) modelled a multi-layer network with feedforward, feedback, and recurrent (local inhibition) connections which were trained with a supervised error-driven learning and tested on a synthetic 3D object (CU3D-100) images dataset (Wyatte et al., 2012; O'Reilly et al., 2013). Their model showed that top-down connections can fill in missing information in partially occluded images and recurrent connections improved the robustness of the model for high levels of occlusion. Sa-Couto et al. (2023) created sparse distributed codes of images (MNIST and F-MNIST dataset) suitable for encoding into a Willshaw network with binary recurrent weights and use it storing and recalling images (Sa-Couto and Wichert, 2020; Sacouto and Wichert, 2023; Simas et al., 2023b). Salvatori et al. (2021, 2024) and Tang et al. (2023) focused on combining predictive coding models aimed at associative memory tasks. The learning process is governed by a covariance-based predictive coding rule termed covPCN, which explicitly encodes the precision (covariance) matrix. Ravichandran et al., (2023) used projections employing Hebbian-Bayesian learning with structural plasticity for feedforward and recurrent projections to create associative memories and showed recurrence improved the robustness of the model to distortions of various kind (Ravichandran et al., 2023a). Traditional deep learning models, such as convolutional neural networks, have been augmented with recurrent connections (Hopfield network) in their penultimate layer to improve their robustness as well as to capture neural dynamics of the mammalian visual cortex and behavioral performance (Tang et al., 2018; Kar et al., 2019; Kietzmann et al., 2019b; Kar and DiCarlo, 2021).

## 3  Model description

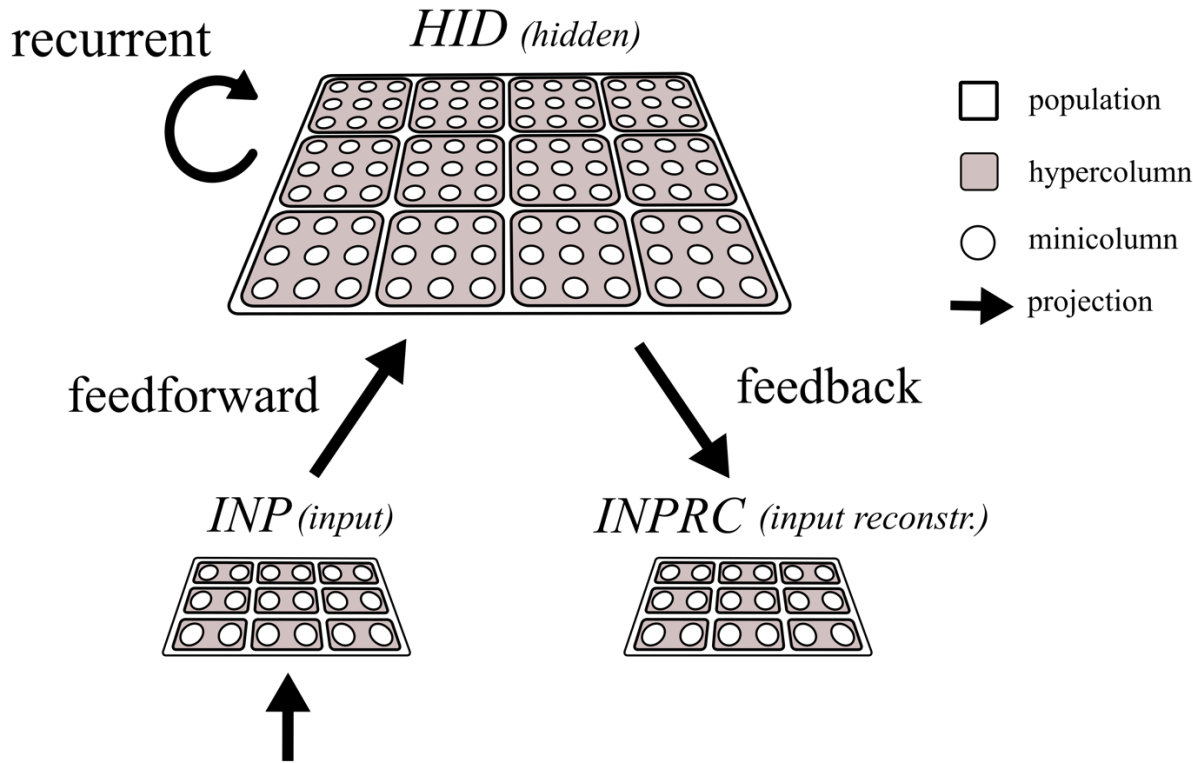

*Figure 2. Schematic of the network architecture. The input (INP), hidden (HID), and input reconstruction (INPRC) populations follow the columnar/modular architecture, i.e. they are modularized into hypercolumns, which in turn constitute minicolumn units locally competing through softmax normalization. The feedforward projections connect the INP with the HID populations, recurrent projections connect the HID units with other HID units, and the feedback projections connect the HID population with the INP population.*

### 3.1  Architecture

The network constitutes three populations, *INP*, *HID*, and *INPRC* where the *INP* population is connected to the *HID* population with a feedforward projection to perform representation learning. The units in the *HID* population are recurrently connected to perform associative memory function, and the *INPRC* population receives feedback projection from the *HID* population for reconstructing the inputs.

Each population is modularized into hypercolumns and minicolumns following the columnar organization of the mammalian neocortex (Mountcastle, 1957, 1997; Hubel and Wiesel, 1962; Fransen and Lansner, 1998; Douglas and Martin, 2004; Bastos et al., 2012). The brain's cortical minicolumn comprises around 80-100 tightly interconnected neurons having functionally similar response properties (Hubel and Wiesel, 1962; Buxhoeveden and Casanova, 2002) and we abstract them into a single functional unit in this work. The minicolumn units (shown as white circles in Fig. 2) locally compete within the hypercolumn module (shown as filled squares enclosing the white circles in Fig. 2) operationally defined as the extent of local lateral inhibition. Thus, each population is composed of several identical hypercolumn modules, each of which in turn comprises many minicolumn units.

## 3.2 Hebbian-Bayesian learning

The learning rule changes the synaptic strength of connections using Bayesian-Hebbian synaptic plasticity. Our learning rule makes use of the local information available spatiotemporally at the synapse, making the learning mechanism and its underlying synaptic plasticity Hebbian, localized, and online. We indicate the pre- and post-synaptic population for each projection with the subscript $i$ and $j$, respectively. From the pre- and post-synaptic spike trains, $s_i, s_j \in \{0, 1\}$, the learning rule involves calculating a cascade of terms accumulating the short- and long-term statistics of pre-, post-, and pre-post joint spiking activity. All the spike and trace variables are time dependent (time index is dropped for notation brevity).

The $z$-traces compute the short-term filtered signals from the pre- and post-synaptic spikes (Equation 1). The $z_i$-trace is modelled after the rapid calcium influx initiated by opening of the NMDA and AMPA channels with short time-constants ($\tau_{z_i} \approx$ 5–100 ms). The $z_j$-trace is modelled after the post-synaptic depolarization event and backpropagating action potential time-constants ($\tau_{z_j} \approx$ 5–100 ms). Each spike event is scaled by a spike scaling factor, $\mu_{spk} = f_{max} \Delta t$, where $f_{max}$ is the hyperparameter controlling the maximal firing-rate and $\Delta t$ is the timestep ($\Delta t$ = 1 ms).

$$\tau_{z_i} \frac{dz_i}{dt} = \frac{1}{\mu_{spk}} s_i - z_i, \qquad \tau_{z_j} \frac{dz_j}{dt} = \frac{1}{\mu_{spk}} s_j - z_j \qquad (1)$$

The $z$-traces provide the coincidence detection window between pre- and post-synaptic spikes for subsequent plasticity induction. The $z$-traces are further transformed into $p$-traces, $p_i, p_j$, and $p_{ij}$, with long time-constants $\tau_p$ (seconds to hours) reflecting the long-term synaptic plasticity process.

$$\tau_p \frac{dp_i}{dt} = z_i - p_i, \qquad \tau_p \frac{dp_{ij}}{dt} = z_i z_j - p_{ij}, \qquad \tau_p \frac{dp_j}{dt} = z_j - p_j \qquad (2)$$

The $p$-traces are finally transformed to bias and weight parameters of the synapse corresponding to terms in classical artificial neural networks. The bias term represents the self-information (or surprisal, or log prior) of the post-synaptic minicolumn unit, and the weight term – the point-wise mutual information between pre- and post-synaptic minicolumn units:

$$b_j = \log p_j, \qquad w_{ij} = \log \frac{p_{ij}}{p_i p_j} \qquad (3)$$

As a crucial departure from traditional backprop based DNNs, the learning rule above is local, correlative, and Hebbian, i.e., dependent only on pre- and post-synaptic activities.

Synaptic plasticity is grounded in the BCPNN framework, which integrates probabilistic inference into biologically plausible neural and synaptic operations. Our previous work with rate-based models showed that this learning rule is equivalent to the expectation maximization algorithm on a discrete mixture model where each minicolumn codes for a discrete mixture component (Ravichandran et al., 2021, 2024).

### 3.3 Spiking activation

The population in our network constitutes neurons producing spiking output where the firing rate reflects the confidence, i.e., probability of the presence of feature conditioned on the pre-synaptic population activity. The total synaptic input for neuron $j$ is updated to be the weighted sum of incoming filtered spikes. The membrane voltage, $v_j$, is updated as the total synaptic input with a time constant $\tau_m$ as follows:

$$\tau_m \frac{dv_j}{dt} = b_j + \sum_{i=0}^{N_i} z_i \, w_{ij} \, c_{ij} + I_j^{ext} - v_j, \quad (4)$$

where $I_j^{ext}$ denotes the external current input and $c_{ij}$ is the binary connection variable, $c_{ij} \in \{0,1\}$, indicating the presence of an *active* or *silent* connection (learned by the structural plasticity mechanism described in Section 3.4). The spiking probability of the neuron $j$, $\pi_j$, is computed as a softmax function over the membrane voltage, which induces a soft-winner-takes-all competition (lateral inhibition) between the neurons within the hypercolumn module. The output of the softmax function reflects the posterior belief probability of the minicolumn unit according to the BCPNN formalism.

$$\pi_j = \frac{\exp(v_j)}{\sum_{j'=1}^{M_j} \exp(v_{j'})}, \quad (5)$$

In the non-spiking (rate based) BCPNN model, this activation $\pi_j$ acts as the firing rate and can be directly communicated as the neuronal signal. For our spiking model, we formulate the instantaneous firing rate as the posterior belief probability ($\pi_j$) scaled by the spike scaling factor ($\mu_{spk}$) and draw independent binary samples, $s_j$, with a spike probability (event with value of 1) as follows:

$$s_j \sim P(\text{spike between t and t} + \Delta t) = \pi_j \, \mu_{spk} \quad (6)$$

For timestep $\Delta t$ smaller than the duration of changes in the underlying firing rate, the spike sampling process approximates the discrete-time version of the Poisson distribution with the underlying firing rate acting as the Poisson mean $\lambda$ (Buesing et al., 2011; Ravichandran et al., 2023c). The scaling factor ($\mu_{spk} = f_{max} \Delta t$) scales the posterior belief probability to the maximum firing rate set by $f_{max}$ (< 1000 Hz) and this renders the filtered spike statistics of the model to be equivalent to the rate-based model (Ravichandran et al., 2023c).

### 3.4 Structural plasticity for network rewiring

Our structural plasticity algorithm (Ravichandran et al., 2020) corresponds to the concept of structural plasticity in the brain which removes existing synaptic connections and creates new ones, thereby modifying the structure of the network in an activity- and experience-dependent manner (Bailey and Kandel, 1993; Lamprecht and LeDoux, 2004; Stettler et al., 2006; Butz et al., 2009; Holtmaat and Svoboda, 2009). Based on the current knowledge about neocortical circuits, we incorporated three key experimental findings in our algorithm: (1) the number of synaptic contacts (incoming connections) made on pyramidal (excitatory) neurons remains roughly constant throughout the neocortex, (2) neocortical connectivity is highly patchy, i.e., axons originating from pyramidal neurons branch a few times and terminate in local spatial clusters making thousands of synapses with spatial extent of the same order as a hypercolumn, and (3) many of the synaptic contacts made on the pyramidal neurons are "silent", i.e., synapses which are physical present but do not allow synaptic transmission.

Based on these observations, our rewiring algorithm computes for each *active* connection patch between every sending and receiving hypercolumn a normalized mutual information score ($\widetilde{M}$). The $\widetilde{M}$ score between each sending and receiving hypercolumn is defined as follows:

$$\widetilde{M} = \frac{\sum_{i,j} p_{ij}\, w_{ij}}{\sum_j c_{ij}}, \qquad (7)$$

where the indices $i$ and $j$ are minicolumn indices summing within their respective hypercolumns. The numerator is equivalent to the mutual information computed locally available at each connection patch ($w_{ij}$ are point wise mutual information as described in Eq. 3) and the denominator is the number of outgoing connections per sending hypercolumn. For each receiving hypercolumn, if some *silent* incoming connection has greater score than some incoming *active* connection, their roles are flipped so that the *active* connection becomes *silent* and vice versa. The *silent* connections have zero weight but still act as "Hebbian probes" for statistics corresponding to the use of *silent*/*active* synapses in the (Isaac et al., 1995; Liao et al., 2001; Kerchner and Nicoll, 2008) and modeling literature (Stepanyants et al., 2002; Knoblauch and Sommer, 2016). The $\widetilde{M}$ score is maximized by each receiving hypercolumn by performing flip operations where we define a flip operation as follows: converting a *silent* connection with the highest $\widetilde{M}$ score into an *active* connection and converting an *active* connection with the lowest $\widetilde{M}$ score to a *silent* connection. We perform $N_{flip}^{conn} = 100$ for every step of structural plasticity and we perform structural plasticity step once for every $N_{intv}^{conn} = 200$ training patterns. This way, the rewiring algorithm operates on the connectivity matrix of each projection and uses the locally available statistics on each connection patch to learn a sparse patchy connectivity.

## 4    Experimental setup

### 4.1    Core dataset

In this work, we used the MNIST hand-written digits dataset (LeCun et al., 1998), a popular image recognition benchmark dataset in the machine learning domain (accessible at http://yann.lecun.com/exdb/mnist/). MNIST consists of 60000 training images and 10000 test images, each with an image and the associative label denoting one of the ten classes. The MNIST images are 28x28 pixel grayscale images with one digit per image. The pixel values are grayscale intensities indicating ink stroke (1 for ink and 0 for blank space), which can be interpreted as the probability of pixel being turned on while feeding into our network. The class labels were not used for training our network model.

## 4.2 Test data for three associative memory tasks

We derived three distinct test image datasets from the MNIST dataset to evaluate performance in three associative memory tasks, namely pattern completion, perceptual rivalry, and distortion resistance. In each case we used the first 1000 samples of the MNIST test dataset (for testing the model, not training). We varied the difficulty level of each task using the "difficulty level" ∈ {0.2, 0.4, 0.6, 0.8, 1}, and for each of the five difficulty levels, we created 1000 patterns making it 5000 patterns in total for each task (what "difficulty" implies for the tasks varies in each case and we describe it in detail below).

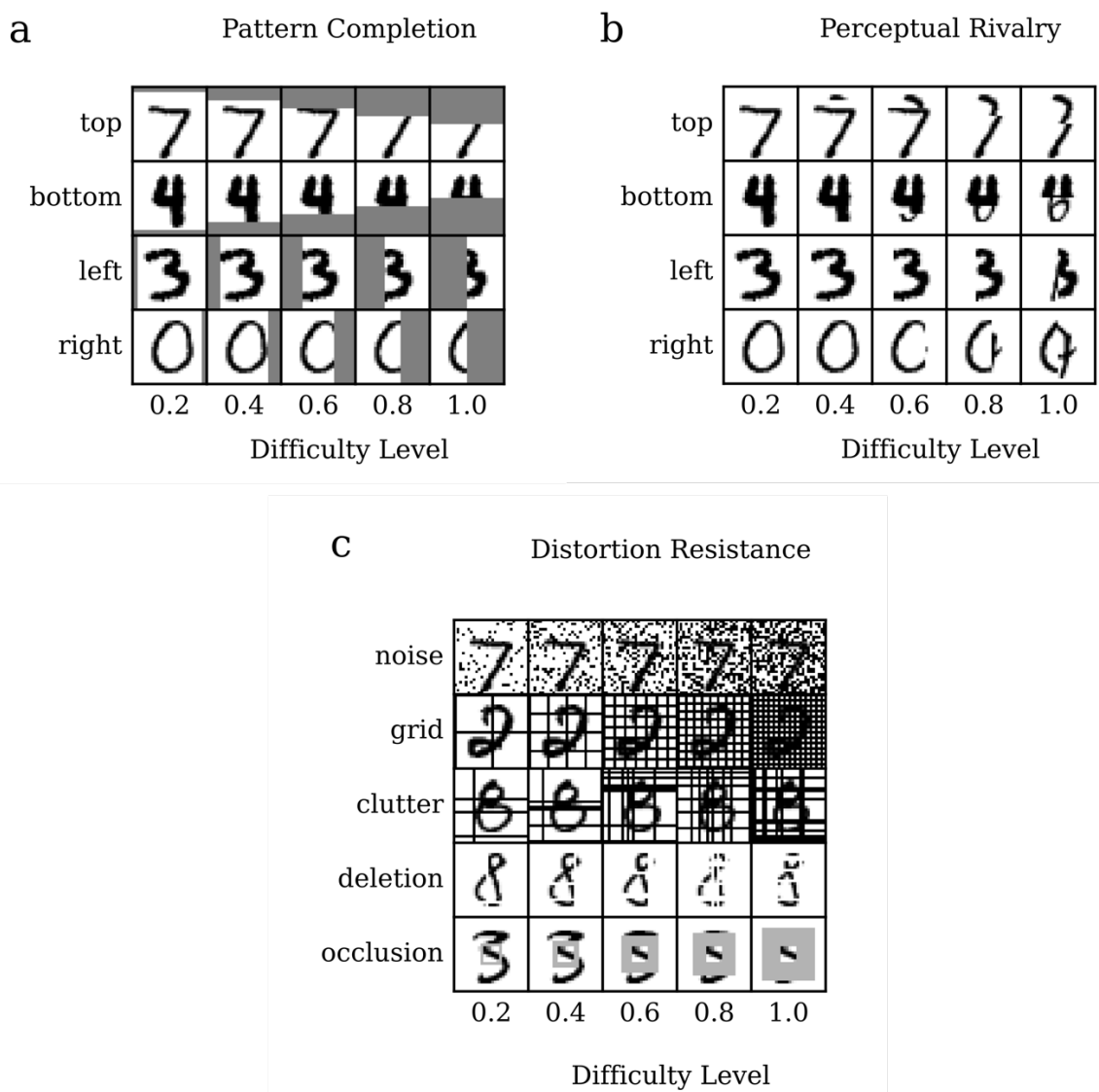

*Figure 3. MNIST dataset modified for pattern completion, perceptual rivalry, and distortion resistance tasks sorted by difficulty level (column-wise) and type of modification (row-wise). (a) Pattern*

*completion: The images are partially visible as a gray bar of varying width is placed on either the top, bottom, left, or right of image. (b) Perceptual rivalry: The images are partially overlapped with varying width by a randomly chosen rival image. (c) Distortion resistance: The images are modified by adding random flips (noise), regularly spaced grid lines (grid), randomly spaced black lines (clutter), randomly spaced white lines (deletion), or a grey square box (occlusion) with varying degree controlled by the difficulty level.*

For the pattern completion task, the associative memory model was expected to recover the original memory pattern when presented with partial patterns. To simulate this, we modified the MNIST images by placing a gray bar of varying width and varying position on the image (for examples, see Fig. 3a). The bar had pixel intensities of 0.5, interpreted as turning on the given pixel at a chance level (c.f. Eqs. 4 and 5). For each difficulty level, $D$, the width of the bar (in pixels) was computed as follows: $width = 14 * D$ (14 is half the size of MNIST image). We chose four positions for the placement of the bars, up, down, left and, right, each amounting to 250 patterns per difficulty. For the pattern rivalry task, the associative memory model was presented with multiple conflicting patterns (typically two), and the model was expected to render one pattern to "win over" the others rival patterns. To simulate this, we modified the MNIST images by replacing a bar of varying width with pixels from another image (for examples, see Fig. 3b).

For each difficulty level, $D$, the width of the rival image was calculated as $width = 14 * D$. We choose four positions for the placement of the bars, up, down, left and, right, each amounting to 250 patterns per difficulty. The rival images were chosen pseudo-randomly by progressing within the 250 test patterns in the reverse direction, for e.g., the 8th image had the 242nd image as the rival (this rendered the procedure deterministic for simplicity).

For the distortion resistance task, the associative memory model was presented with patterns under various distortions and the model was supposed to restore the original, undistorted ones. To simulate this, we modified the MNIST images by performing one of five types of distortion (for examples, see Fig. 3c). For each difficulty level, $D$, we split the 1000 test images into 5 distortion types and created the following five distortions to the images: noise, grid, clutter, deletion and, occlusion, derived from previous work (George et al., 2017; Ravichandran et al., 2023a).

### 4.3 Network setup

The *INP* population constituted $H_{INP} = 784$ hypercolumns corresponding to pixels of MNIST flattened 28x28 image. Each hypercolumn was made up of $M_{INP} = 2$ minicolumns corresponding to the binary nature of the pixel intensity (ON or OFF). The *HID* population constituted $H_{HID} = 100$ hypercolumns with $M_{HID} = 100$ minicolumns per hypercolumn. The input reconstruction population, *INPRC*, had the same shape as the *INP* population, with $H_{INPRC} = 784$ and $M_{INPRC} = 2$. For all the projection types we set the parameter $N_{conn}$ which determines the number of incoming connections per receiving hypercolumn (these connections were inherited by all minicolumns units within any given hypercolumn).

### 4.4 Simulation protocol

The MNIST image was injected as input into the network by setting the external current of the *INP* and *INPRC* populations. We applied the log of the image pixel intensities as the input current. Since all the populations in our network use the softmax activation function, the logged pixel intensities get transformed to the pixel intensities in the range [0,1] (minicolumn activities). We also clipped the input current to a small positive value (1e-10), which prevents the exponential term in the softmax from becoming negative infinity (they get clipped to -10). For this, the image pixel intensity (indexed by $j$) is injected into the two minicolumns (indexed by $2j$ and $2j + 1$) of the $j$-th hypercolumn as follows:

$$I_{2j}^{ext} = \log u_j, \tag{9}$$

$$I_{2j+1}^{ext} = \log (1 - u_j), \tag{10}$$

where $u_j$ is the pixel intensity of the image normalized to be in the range [0, 1].

The network was first run in the training mode where the feedforward-driven activities are used by the network for synaptic learning and structural plasticity. This is done for each pattern by updating the $p$-traces (Eq. 2) and computing the weights and biases on the last step of each pattern presentation. The network was then run in the evaluation mode where the test and modified test datasets (for associative memory tasks as described in Section 4.2) were run in succession without any learning.

*Table 1. Network parameters for SpspkFull model*

| Type | Parameter | Value | Description |
|---|---|---|---|
| Network architecture | $H_{INP}$ | 784 | # hypercolumns in input population |
| | $M_{INP}$ | 2 | # minicolumns per hypercolumn in input population |
| | $H_{HID}$ | 100 | # hypercolumns in hidden population |
| | $M_{HID}$ | 100 | # minicolumns per hypercolumn in hidden population |
| | $H_{INPRC}$ | 784 | # hypercolumns in input reconstruction population |
| | $M_{INPRC}$ | 2 | # minicolumns per hypercolumn in input reconstruction population |
| | $N_{conn}^{INP \to HID}$ | 78 | # incoming feedforward connections per hidden hypercolumn |
| | $N_{conn}^{HID \to HID}$ | 100 | # incoming recurrent connections per hidden hypercolumn |
| | $N_{conn}^{HID \to INPRC}$ | 10 | # incoming feedback connections per input hypercolumn |
| | $\tau_m$ | 0.005 s | Membrane time constant |

| Neural and synaptic time-constants | $\tau_{zi}, \tau_{zj}$ | 0.020 s | Time constant of Z-traces |
|---|---|---|---|
| | $\tau_p$ | 5 s | Time constant of P-traces |
| Stimulation protocol | $\Delta t$ | 0.001 s | Simulation timestep |
| | $T_{spk}$ | 0.001 s | Time duration of spike |
| | $T_{no-input}$ | 0.100 s | Time duration of no-input phase |
| | $T_{ffwd}$ | 0.100 s | Time duration of feedforward phase |
| | $T_{overlap}$ | 0.050 s | Time duration of overlap phase |
| | $T_{recr}$ | 0.150 s | Time duration of recurrent phase |
| Data setup | $N_{train}$ | 60000 | Number of training patterns |
| | $N_{test}$ | 10000 | Number of test patterns |
| | $N_{epoch}$ | 20 | Number of training epochs |

For each pattern in the training mode the network was run in two phases, *no-input* and *ffwd* phases, and in the evaluation mode the network was run in four phases, *no-input*, *ffwd*, *overlap*, and *recr*, in succession:

1. *no-input* phase – the network is run without any input in order to clear any previous activity and avoid interference,
2. *ffwd* phase – the network is driven with the external input to the *INP* population, in turn, the *INP* population drives the *HID* population,
3. *overlap* phase – the *HID* population is driven both by the *INP* population and itself through recurrent projections, and
4. *recr* phase – the input is cutoff, and the *HID* population is running solely through recurrent self-projections.

We set the duration of each phase using parameters, $T_{no-input}$, $T_{ffwd}$, $T_{overlap}$, and $T_{recr}$ respectively. For simulating the four phases (illustrated in Fig. 4), we controlled the injection of input into the populations (Eqs. 9 and 10) as well as the propagation of activity through each projection individually (Eq. 4). Table 1 summarizes all the default parameters used in our model for the *SpspkFull* model.

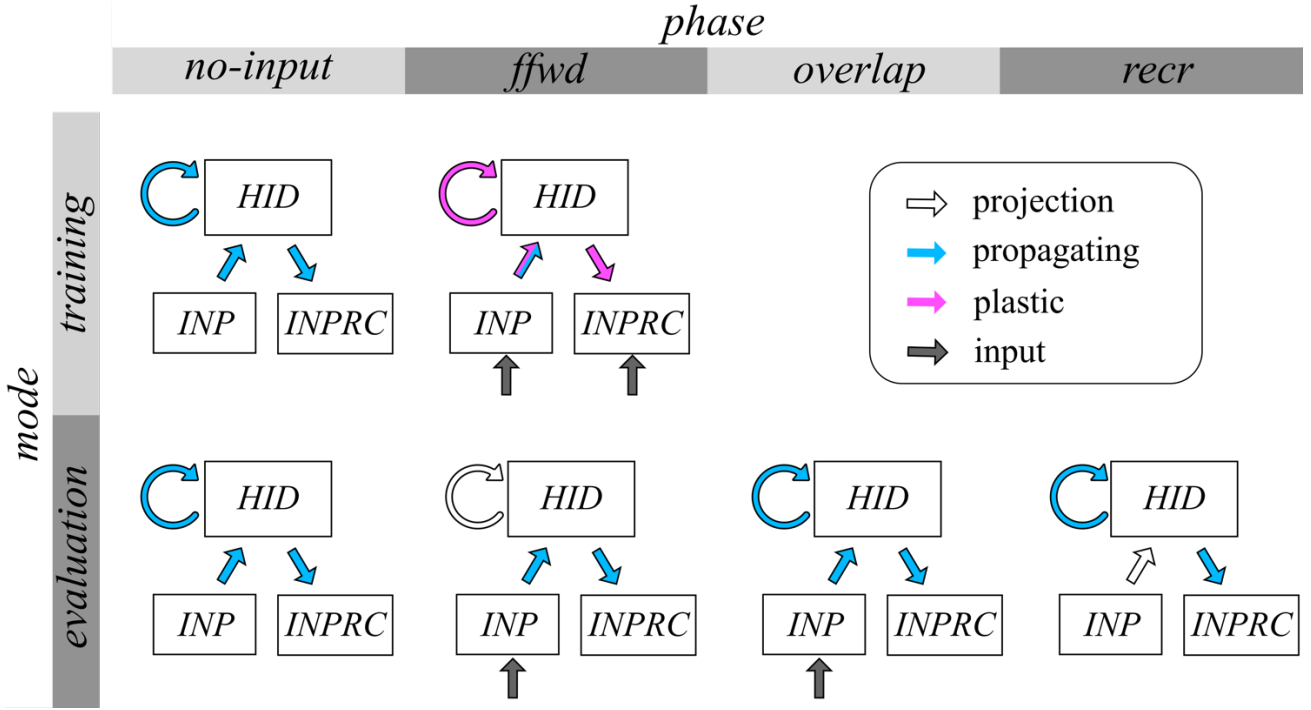

*Figure 4. Protocol for network simulation. The network is run either in training mode (top row), where the feedforward-driven activities are used for synaptic learning and structural plasticity, or in evaluation mode (bottom row), where the feedforward-driven activities are used to cue the recurrence-driven associative memories. For each pattern, the network is run either for two phases (no-input and ffwd) in the training mode or for four phases in the evaluation mode (no-input, ffwd, overlap, and recr). The gray arrows converging on the INP and INPRC populations indicate injecting the MNIST images as inputs to the respective populations. The white unfilled arrows indicate the presence of a projection in the network connecting two populations. Blue filling on the projection arrows indicates propagation of activities through the respective projections. Purple filling on the projection arrows indicate the projection undergoes synaptic and structural plasticity updates for each pattern. Arrow with both blue and purple filling indicates the projection is both propagating and learning. In the no-input phase, the network is run without any input to clear any previous activity. In the ffwd phase the network is driven with the external input to the INP and INPRC populations and the INP population drives HID population. In the overlap phase the HID population is driven both by the INP population and itself through recurrent projections. In the recr phase the input is cutoff, and the HID population is running solely through recurrent self-projections.*

### 4.5 Models under comparison

We compared six different BCPNN models: (1) *RateFf*: rate-based activation in a feedforward network (without recurrent projection), (2) *RateFull*: rate-based activation in a full network (with recurrent projection), (3) *SpkFf*: spiking activation in a feedforward network, (4) *SpkFull*: spiking activation in a full network, (5) *SpspkFf*: sparsely spiking activation (with 100 Hz maximum firing rate) in a feedforward network, and (6) *SpspkFull*: sparsely spiking activation in a full network.

For the rate-based models (*RateFf* and *RateFull*), the activation was implemented by considering the softmax output value (Eq. 5), $\pi_j$, directly as the neuronal signal communicated across the network. For spiking (*SpkFf* and *SpkFull*) and sparsely spiking (*SpspkFf* and *SpspkFull*) activation, we further sampled binary values (Eq. 6), $s_j$, from the softmax output and used these for communication. Crucially, all the six models were simulated in the same code implementation by modifying the parameter values as listed in Table 2.

*Table 2. Parameters for the six models under comparison*

| Parameter | *RateFf* | *RateFull* | *SpkFf* | *SpkFull* | *SpspkFf* | *SpspkFull* |
|---|---|---|---|---|---|---|
| Activity | $\pi_j$ | $\pi_j$ | $s_j$ | $s_j$ | $s_j$ | $s_j$ |
| $f_{max}$ (Hz) | - | - | 1000 | 1000 | 100 | 100 |
| $\tau_{zi}, \tau_{zj}$ (s) | 0.001 | 0.001 | 0.005 | 0.005 | 0.020 | 0.020 |
| $\tau_m$ (s) | 0.001 | 0.001 | 0.001 | 0.001 | 0.005 | 0.005 |
| $T_{no-input}$ (s) | 0 | 0 | 0.025 | 0.025 | 0.100 | 0.100 |
| $T_{ffwd}$ (s) | 0.005 | 0.005 | 0.025 | 0.025 | 0.100 | 0.100 |
| $T_{overlap}$ (s) | 0 | 0 | 0 | 0.025 | 0 | 0.050 |
| $T_{recr}$ (s) | 0 | 0.020 | 0 | 0.050 | 0 | 0.150 |

### 4.6 Evaluating representations via linear classifier

We used a linear classifier trained on the model's internal representations to decode the class labels. Although our model does not require class labels for learning, we exploited the label information to quantify the class separability as one of the evaluation methods. For this, we used a simple linear classifier with *N*=10 softmax output units corresponding to the class labels.

We used the *z*-traces of the hidden population as the input to the classifier. For training the classifier, we used the cross-entropy loss function and Adam optimizer with parameters $\alpha = 0.001$, $\beta_1 = 0.9$, $\beta_2 = 0.999$ and, $\epsilon = 10^{-7}$ as originally defined (Kingma and Ba, 2015). We used minibatches of 64 samples and trained the network for 10 epochs.

## 5 Results

### 5.1 Sparsely firing representations show orthogonalization necessary for associative memory

Associative memory models require the patterns stored to be sparse orthogonal with minimal overlap, so that the attractor memories do not suffer interference or form spurious minima (Amit et al., 1987). We investigated with our *SpspkFull* model if the feedforward-driven activities form sparse spiking representations that have minimal overlap and benefit the formation of robust recurrent-driven associative memories.

To this end, we first visualized the spike raster and firing rate of selected hypercolumns from the *INP*, *HID*, and *INPRC* populations of the *SpspkFull* model (run on evaluation mode). The spike raster of all the populations (Fig. 5; upper row) shows that the activities are mostly silent with occasional brief periods of high frequency bursts. We further computed the firing rates by convolving spike trains with a Gaussian kernel ($\sigma^2$ = 20 ms) (Fig. 5; lower row). We observed that very few minicolumns produced high firing rates (typically one within a hypercolumn module) at any time due to the softmax normalization performed by the model. The peak firing rate of these units typically reaches 100 Hz confirming our model operating based on the scaling operation on the firing probability by the $f_{max}$ parameter ($f_{max}$ = 100 Hz; Eq. 6).

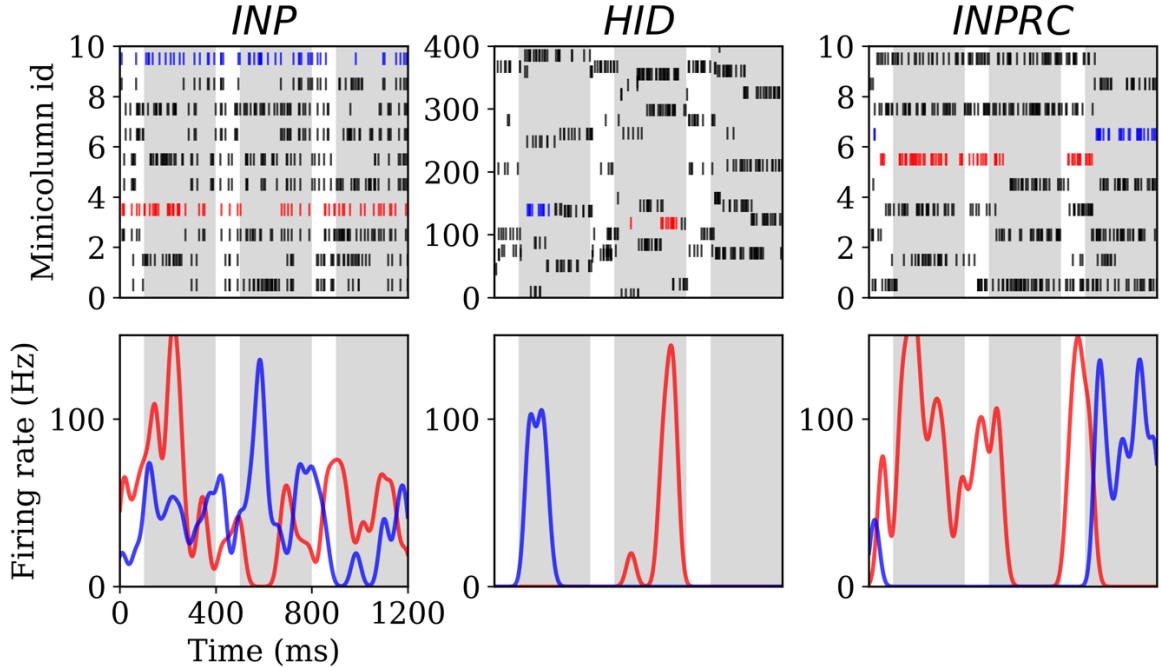

*Figure 5. Spiking activity and firing rate for the INP (left), HID (center), and INPRC (right) populations. The first row shows the spike rasters for the first 10 minicolumns (arranged within 5 hypercolumns) for the INP and INPRC population and the first 400 minicolumns (arranged within 4 hypercolumns) for the HID population. The white bars correspond to the no-input period (100 ms) and the gray bars correspond to the pattern period (ff, overlap, and recr combined; 300 ms). The bottom row shows the firing rate of two selected minicolumns (indicated by blue and red spike trains in their respective population) computed by convolving the spike trains with a Gaussian kernel ($\sigma^2$ = 20 ms) demonstrating the minicolumns that spike with the maximum firing rate around 100 Hz.*

Next, we sought to assess the degree of orthogonalization of the representations by computing the representational similarity of all populations. For this, we computed the pair-wise cosine similarity between the z-traces for test patterns ($N_{test}$ = 10000) at two time points after the pattern onset: $T$ = 100 ms ($T_{ffwd}$; feedforward-driven activities) and $T$ = 300 ms ($T_{ffwd} + T_{overlap} + T_{recr}$; attractor-driven activities). For each of the similarity matrices, we computed the orthogonality ratio, $s_{ortho}$, that measures the ratio between average within-class similarity and average similarity across all pairs of patterns; higher value of $s_{ortho}$ implies higher degree of orthogonalization.

The similarity matrix for *INP* population (Fig. 6; left) shows the within-class similarities (class-wise diagonal values) are not very distinct from the between-class similarities (class-wise off-diagonal values), with $s_{ortho}$=1.04, directly reflecting the nature of the MNIST dataset where images are highly correlated. Any potential associative memory directly trained on the *INP* representations would suffer from severe interference between memories owing to the high degree of overlap. The similarity matrices for *HID* representations at $T = 100$ and $300$ ms (Fig. 6; upper and lower center) are strikingly different from those corresponding to the *INP* representations ($s_{ortho} = 3.67$ at $T = 100$ and $s_{ortho} = 7.01$ at $T = 300$ ms), demonstrating a strong trend for the orthogonalization of *HID* representations: mostly zero similarity between-class while within-class similarity are visibly distinct and high in magnitude. The similarity matrix at $T = 300$ ms is "crisper" compared to the same at $T = 100$ ms, as expected from associative memory dynamics acting on the feedforward-driven representations. The similarity matrix for *INPRC* population at $T = 100$ and $300$ ms (Fig. 6; upper and lower right) shows stronger orthogonalization of the reconstructed representations ($s_{ortho} = 1.07$ at $T = 100$ ms and $s_{ortho} = 1.14$ at $T = 300$ ms) in comparison with the input representations. This demonstrated that the *HID* representations showed high degree of orthogonalization that would benefit the associative memory to store patters without considerable interference when compared to the *INP* representations.

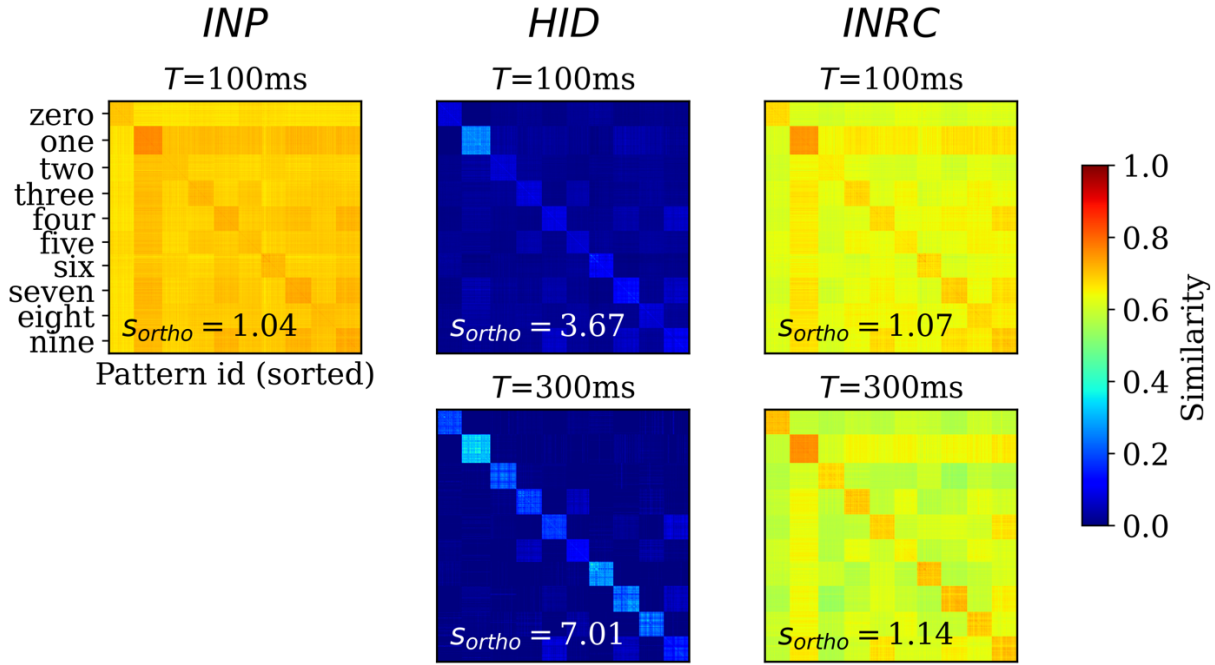

*Figure 6. Representational similarity. Pair-wise cosine similarity matrices for N=10000 MNIST test patterns sorted by their labels for INP (left), HID (middle) and INPRC (right) populations at T = 100 ms (feedforward-driven; upper row) and T = 300 ms (attractor-driven; lower row) representations. The orthogonality ratio, $s_{ortho}$, is displayed inside each plot. The INP population shows low orthogonality due to the large similarity values (0.6-0.8) both within- and between-classes. The HID population (T=100 and 300 ms) shows high orthogonality due to low similarity values (0-0.2) between-*

*class owing to the sparse distributed nature of representations. The orthogonality ratio also increases from feedforward-driven representations (T=100 ms) to attractor representations (T=300 ms). The input reconstruction population, INPRC, shows more orthogonality when compared to the corresponding INP similarities.*

Based on the above experiments, we concluded that the *HID* activities exhibited properties of sparse orthogonal representations that are suitable for associative memory formation. We further qualitatively examined the evolution of *INP* and *INPRC* population outputs (Fig. 7) after stimulating with one example pattern for a time-period of $T = 300$ ms after the pattern onset (we skipped the no-input period). For this, we visualized the raw spiking activities ($s_j$) and the short-term filtered z-traces (for *INP* population we used the $z_i$-traces of feedforward projection and for *INPRC* population – the $z_j$-traces of feedback projection). The *INP* spiking activity shows highly sparse sampling of the input during the first 150 ms (feedforward and overlap phase; $T_{ffwd} = 100$ ms and $T_{overlap} = 50$ ms) and noise in the later phase, 150-300 ms (recurrent phase; $T_{recr} = 50$ ms). The *INPRC* representations reflect that there is an initial reconstruction of the presented digit (60-180 ms) corresponding to the feedforward-driven representations. After the inputs are removed (150-300 ms), the reconstruction settles to a prototypical digit, corresponding to the attractor representations in the *HID* population, which is visible as stochastic samples in the spiking activities and clearly visible from the z-traces. Based on similar experiments with many more patterns (not shown for brevity), we observed similar results where the *INPRC* representations converge to a stable attractor resembling a prototypical digit image.

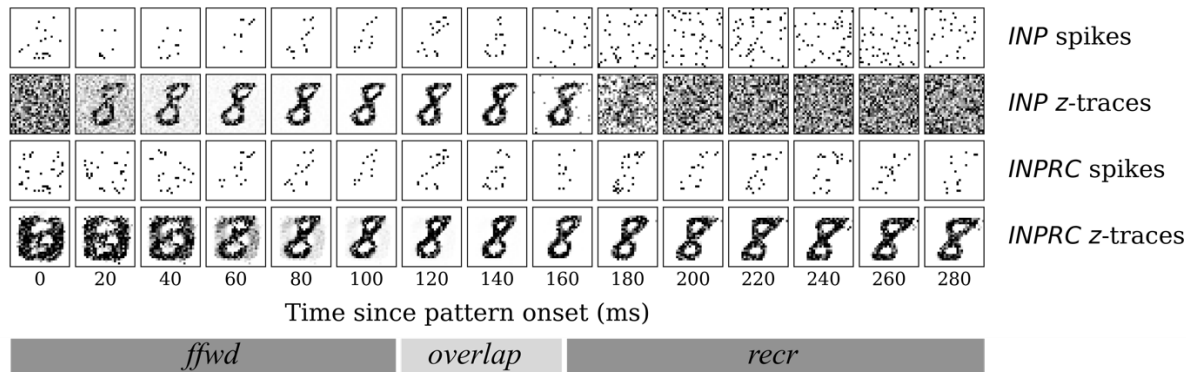

*Figure 7. Time course of attractor representations for one pattern. The spikes and z-traces of INP and INPRC populations for one example pattern for T = 300 ms. The spike raster is highly noisy and sparse while the z-traces show a highly stable representation of digits. The INPRC population shows the initial reconstruction of feedforward-driven representations (T = 60-180 ms) and the attractor reconstructions (T = 150-300 ms) driven by the recurrent projections showing a stable convergence to the prototypical digit (corresponding to one of the class labels) even after the input is no longer fed into the network.*

## 5.2 Structural plasticity forms stable localized receptive fields

We evaluated the effect of the structural plasticity algorithm on the network connectivity and the formation of receptive fields over the course of training. Since the structural plasticity algorithm acts on patchy connections between hypercolumn pairs, we computed the receptive field of each hypercolumn module by using the connectivity matrix of dimensions ($H_{HID}$, $H_{INP}$) for the feedforward projections, and the transpose of the connectivity matrix of dimension ($H_{INP}$, $H_{HID}$) for the feedback projections. We plotted the receptive fields of the first ten *HID* hypercolumns in log steps over the course of training for the feedforward (Fig. 8; left) and feedback (Fig. 8; right) projections.

Each hidden hypercolumn is initialized with randomized connections with the *INP* population (Fig. 8 top rows). The connections converge within the first 10000 training patterns and remain stable over the whole course of training (bottom rows). Moreover, the connections converge to a meaningful set of spatially localized receptive fields over the input space, even though no knowledge of the input space topology was explicitly provided to the network. Furthermore, the receptive fields of the feedforward and feedback projections mirror each other for each hypercolumn module (for instance, first column of left plot and first column of right plot in Fig. 8), demonstrating again that the structural plasticity finds the correlative structure between the *INP* and *HID* populations.

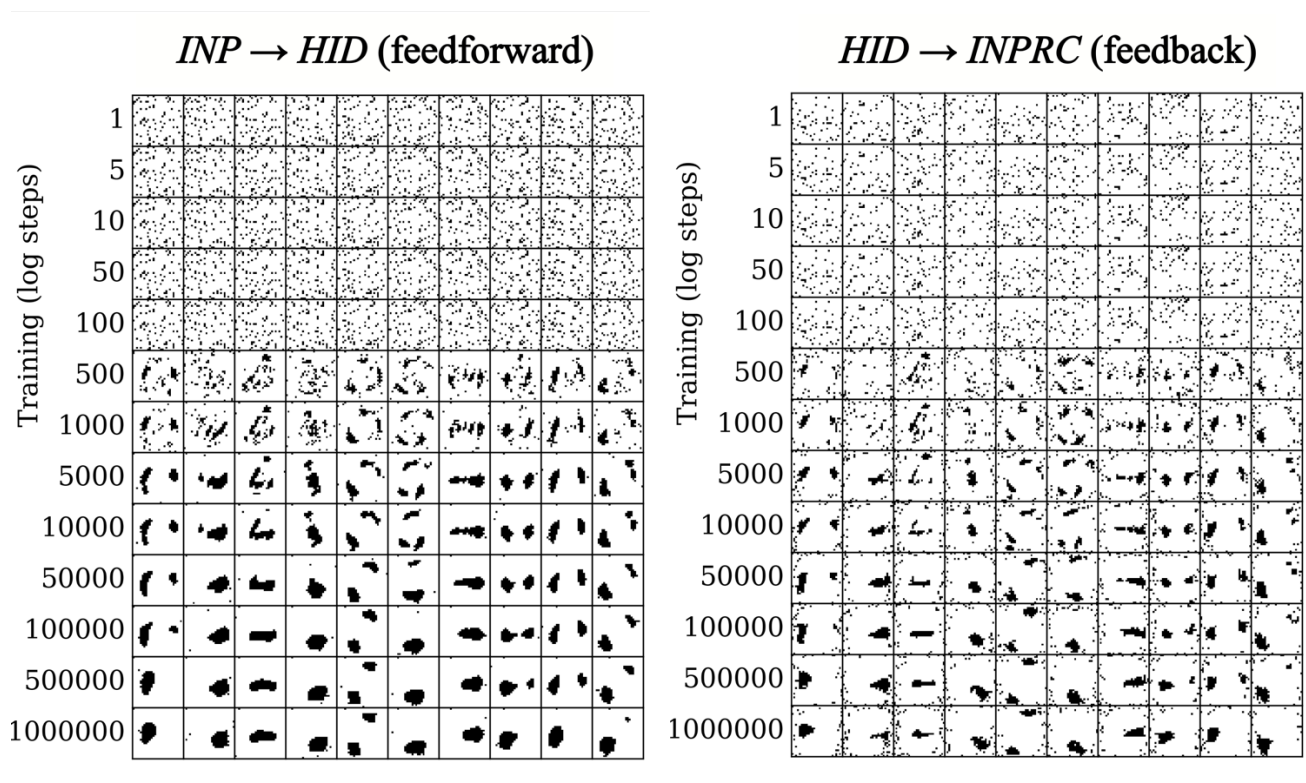

*Figure 8. Receptive field formation for feedforward (left) and feedback (right) projections. Each column corresponds to connections between one randomly chosen hypercolumn of the HID population*

*(column number corresponds to index of HID hypercolumn) and the INP population. Over the course of training the connections form spatially localized receptive fields in the image space.*

We observed that the number of rewiring flip operations for the feedforward and feedback projections (Fig. 9; upper left and right respectively) starts off with a high value at the beginning of the training and decreases over the course of training, converging near zero. The average $\tilde{M}$ score (normalized mutual information), which is greedily maximized by each hidden hypercolumn individually, converges at a high value for both projections (Fig. 9; lower left and right). It is worth noting though that the score has high variability across hypercolumns for the feedback projection. The above results demonstrated the structural plasticity algorithm identifies a set of meaningful localized receptive fields spanning the input space.

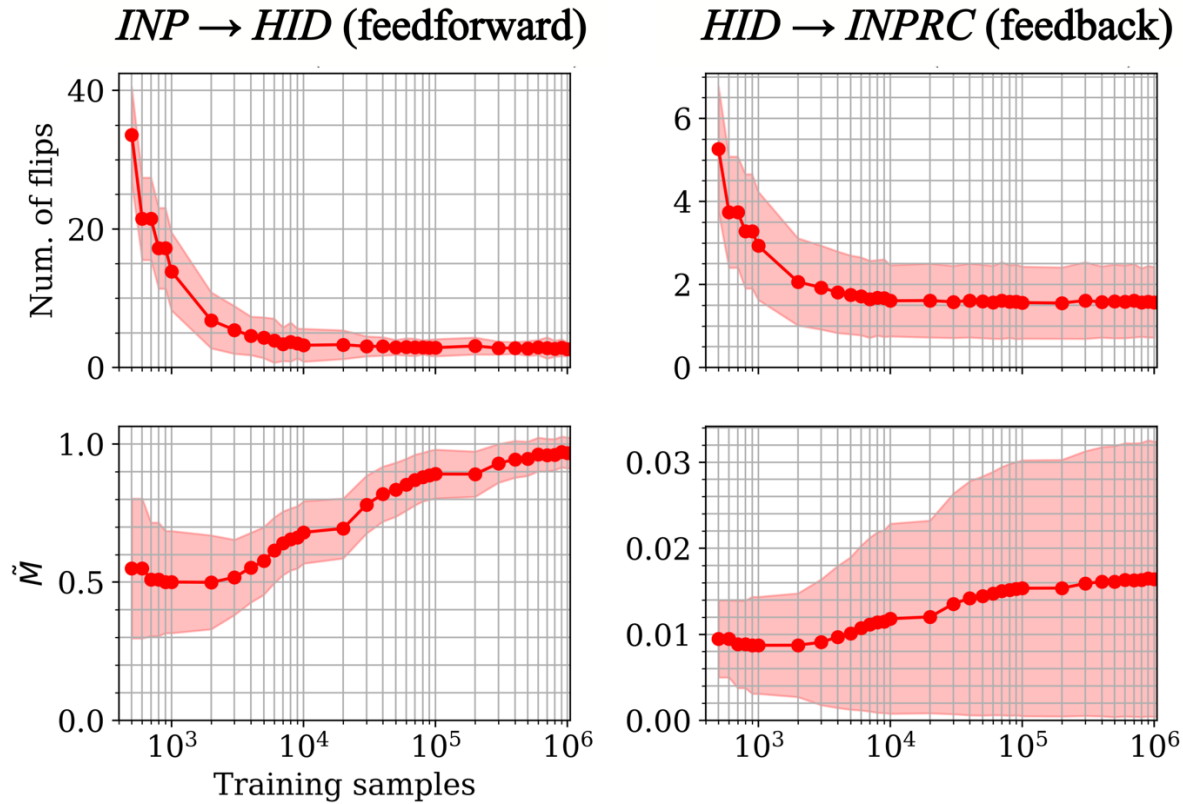

*Figure 9. Convergence of the structural plasticity algorithm for feedforward (left) and feedback (right) projections. The number of rewiring flip operations (top row) and $\tilde{M}$ score (normalized mutual information; bottom row) per rewiring step over the course of training (mean ± std from n=100 HID hypercolumns) shows convergence for both feedforward and feedback projections.*

### 5.3 Short-term *z*-filtering is essential for sparsely spiking networks

The sparsely spiking models (*SpspkFf* and *SpspkFull*) were designed by scaling down the firing rates to low biologically realistic values using the parameter $f_{max}$ (Eq. 7), for instance $f_{max}$ = 100 Hz. The crucial difference between our sparsely spiking models and other models is the use of filtering using high values for time constants $\tau_z$ (short-term *z*-filtering of the pre- and post-synaptic spikes) and $\tau_m$ (membrane time constant). The rate-based models (*RateFf* and *RateFull*) used $\tau_z = 1$ ms and $\tau_m = 1$ ms (= $\Delta t$) which effectively amounts to no filtering and the spiking models (*SpkFf* and *SpkFull*) used relatively low values: $\tau_z = 5$ ms and $\tau_m = 5$ ms.

For the sparsely spiking models, we hypothesized that the effects of scaling down the spiking probability ($f_{max} < 1000$ Hz) can be countered using high values for $\tau_z$ and $\tau_m$ parameters. Furthermore, we expected the maximum firing rate $f_{max}$ and filtering time constants ($\tau_z$ and/or $\tau_m$) to be inversely related, i.e., lower spiking probability (lower $f_{max}$) would be compensated by longer filtering (higher $\tau_z$ and/or $\tau_m$) in order to recapitulate the performance of the spiking model (dense spiking; $f_{max} = 1000$ Hz). To this end, we assessed the *SpspkFull* model performance by systematically varying $f_{max} = \{20, 50, 100, 200, 500, 1000\}$ (in Hz), $\tau_m \in \{1, 2, 5, 10, 20\}$ (in ms), and $\tau_z \in \{1, 2, 5, 10, 20, 50, 100\}$ (in ms) while measuring the linear classification accuracy of each model. Since this experiment involved running many simulations ($n = 210$), we trained the models on a reduced MNIST dataset with $N_{train}=1000$ and $N_{test}=1000$, unlike the rest of the experiments which were trained and tested on the full MNIST dataset with $N_{train}=60000$ and $N_{test}=10000$.

For unrealistically high firing rates ($f_{max} = 200$-$1000$ Hz) we observed the high classification accuracy over a wide range of $\tau_m$ and $\tau_z$ since the spikes are dense samples of the underlying firing rate and filtering is not necessarily helpful (Fig. 10; upper row). For biologically realistic firing rates ($\tau_z = 20$-$100$ Hz) performance with $\tau_z < 10$ ms is very low. This is because pre- and post- synaptic spikes are expected to coincide within this time-window for learning to occur, while the spikes are generated sparsely and irregularly from a Poisson distribution. However, for $\tau_z = 20$-$50$ ms the performance closely approximates the densely spiking model since this time window is sufficient to expect pre- and post-synaptic spikes to coincide and be associated through Hebbian plasticity.

All the runs irrespective of $f_{max}$ drop sharply in performance at $\tau_z = 100$ ms, because the time window provided is too long compared to the presentation time of each pattern ($T = 300$ ms) and learning wrongly associates the current pattern with temporally adjacent patterns. We found the there was no dependence between the value of $\tau_m$ and performance across all values of $f_{max}$. Unlike the z-traces controlling the Hebbian time window of coincidence of pre- and post-synaptic spikes, the membrane time constant $\tau_m$ acts only on the post-synaptic spike and has less significance in the learning phase. The results demonstrated the importance of $\tau_z$ in the functioning of the sparsely spiking models and how longer z-filtering can compensate for the low firing rates. Based on the results, we used $\tau_z = 20$ ms for our sparse spiking models with $f_{max} = 100$ Hz.

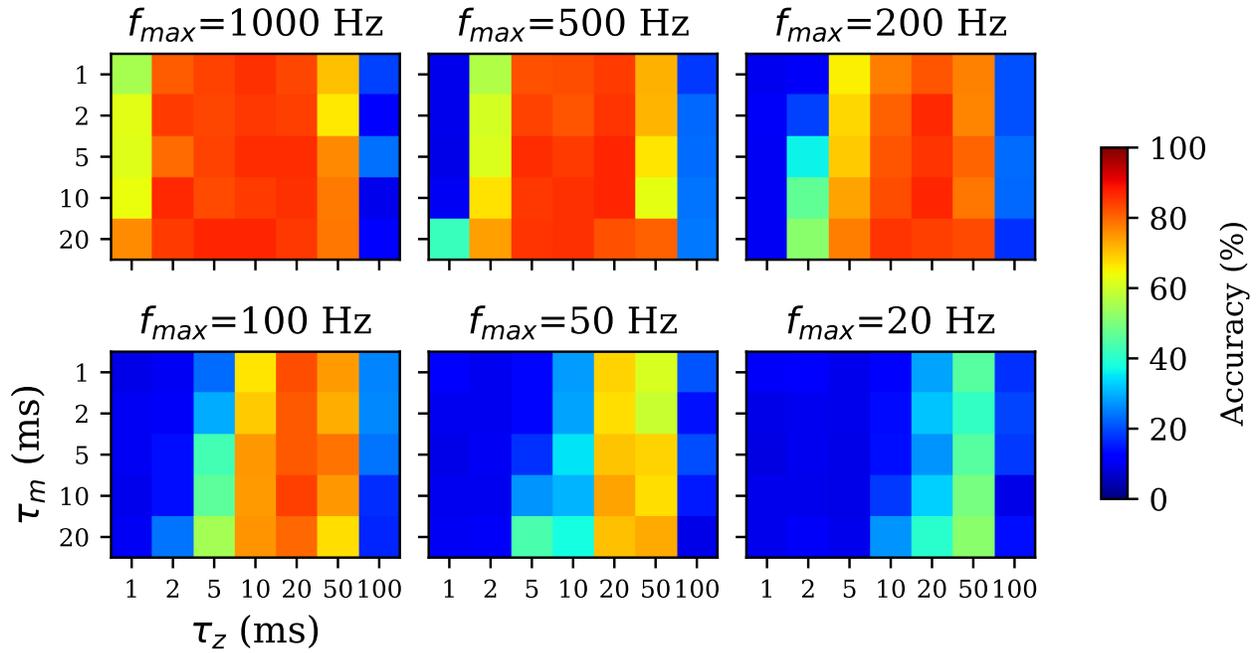

*Figure 10. Impact of z-filtering on classification performance. Higher value of $\tau_z$ and $\tau_m$ implies longer filtering and setting the value to 1 ms (=Δt) implies essentially no filtering. Longer z-filtering compensates for low firing rates in sparsely firing spiking networks. For high firing rate ($f_{max}$>200Hz; top rows), the accuracy is high over a wide range of $\tau_z$ and $\tau_m$ values. For sparsely firing networks with biologically realistic firing rates ($f_{max}$<200Hz; bottom rows), the performance is sensitive to $\tau_z$, optimal in range of 10-50 ms, and resistant to $\tau_m$ values.*

We finally trained all six models (parameter values are listed in Table 2) on the full MNIST dataset ($N_{train}$ = 60000 and $N_{test}$ = 10000) for *n*=5 runs. The test accuracy results (Table 3) show that the *SpspkFf* model closely approximates the performance of *SpkFf* and *RateFf* models. Similarity, *SpspkFull* approximates *SpkFull* and *RateFull* though we notice a small decrease in performance (around 3%). In all cases, the feedforward models (*RateFf*, *SpkFf*, and *SpspkFf*) outperform the full models which feature additional recurrent projections (*RateFull*, *SpkFull*, and *SpspkFull*). This is due to the associative memory changing the feedforward-driven representations into attractor representations. which can occasionally converge to wrong attractors. We discuss this in detail in Section 5.5 and provide scenarios where full models with recurrent projections prove to be beneficial.

Table 3. Test classification performance of all six models on MNIST test dataset (mean ± std % from *n*=5 runs).

|  | *RateFf* | *RateFull* | *SpkFf* | *SpkFull* | *SpspkFf* | *SpspkFull* |
|---|---|---|---|---|---|---|
| Test acc. (%) | 98.06± 0.07 | 95.59±0.14 | 97.93±0.08 | 95.02±0.10 | 97.2±0.08 | 92.38±0.17 |

## 5.4 Prototype extraction

Associative memory involves grouping similar patterns into representative memory objects for storage. Each resulting memory object acts therefore as a representative prototype of the grouped patterns. This is analogous to clustering in machine learning. The prototypes are coded as high-dimensional distributed representations converging to attractor states (Lansner et al., 2023; Ravichandran et al., 2023a).

From the *SpspkFull* model, we expected the recurrent projections to implement prototype extraction and group similar feedforward-driven activities into common attractors. Since the activities in the *HID* population are sparse stochastic spikes, we used the z-traces to measure the pair-wise similarities across all the MNIST test dataset. For this we used the z-traces of the *HID* population at the last step of each pattern run ($T = 300$ ms) and computed the cosine similarity (denoted by $S$) removing the diagonal elements. The distribution of the cosine similarity was heavily skewed towards zero (due to the highly sparse nature of the representations) with a small fraction having a large positive value (above 0.1, for instance; Fig. 11a). We also observed there were no two attractor patterns in the *HID* population that converged on the same unique attractor, as seen by the zero count for $S = 1$. Due to this, we used a threshold value on the cosine similarity, denoted by $S_{min}$, and considered all attractors with similarity above $S_{min}$ as unique prototypes. Since the number of such prototypes found by our method depends on the value of $S_{min}$, we varied $S_{min}$ from 0 to 1 at a regular interval of 0.01 and established the relationship between number of prototypes found and $S_{min}$ (Fig. 11b). We observed that for small values, $S_{min} < 0.25$, there were fewer number of prototypes found (less than 100). For larger values, $S_{min} = 0.75$, the number of prototypes quicky approached the number of total test patterns ($N_{test} = 1000$), since there were very few attractor patterns that had a cosine similarity $S > 0.75$ (Fig. 11a) and almost all attractor patterns were categorized as unique attractors.

We selected three values of similarity threshold, $S_{min} = 0.01, 0.1$, and $0.2$, and examined the prototypes found by the model based on the input reconstructions from the *INPRC* population (Fig. 11c). Since this involved many patterns from the test dataset converging on the same prototype, we averaged all the *INPRC* z-traces (from the last time step, $T = 300$ ms) that were categorized as the same prototype based on the similarity threshold. Hence, we found the average input reconstruction per prototype as well as a "attraction" index that indicates how many patterns converged on the same prototype (displayed as small text at the top of each image in Fig. 11c). For $S_{min} = 0.01$, there were 13 prototypes found that resembled most of the unique digits from the dataset (Fig. 11c; upper row). For $S_{min} = 0.1$, there were 33 prototypes found that covered all the unique digits from the dataset, as well as capturing different styles of writing the same digit, for instance, upright and slanted one (Fig. 11c; middle row). For $S_{min} = 0.2$, there were 65 prototypes found that similarly captured most of the writing styles of digits. We found a highly skewed distribution of the attraction index with only a few prototypes being popular and the majority of prototypes having a small attraction index. We also observed that the input reconstruction of the prototypes is highly stable, i.e., the average input reconstruction of a large number of attractors lead to crisp images. For instance, for $S_{min} = 0.1$, for the three most popular prototypes the average input reconstructions of around 600 patterns clearly resembled digits "1", "8", and "4". Considering that the cosine similarity in the *HID* attractor space was a small value of $S_{min} = 0.1$, the image reconstructions were highly similar to each other in the input space.

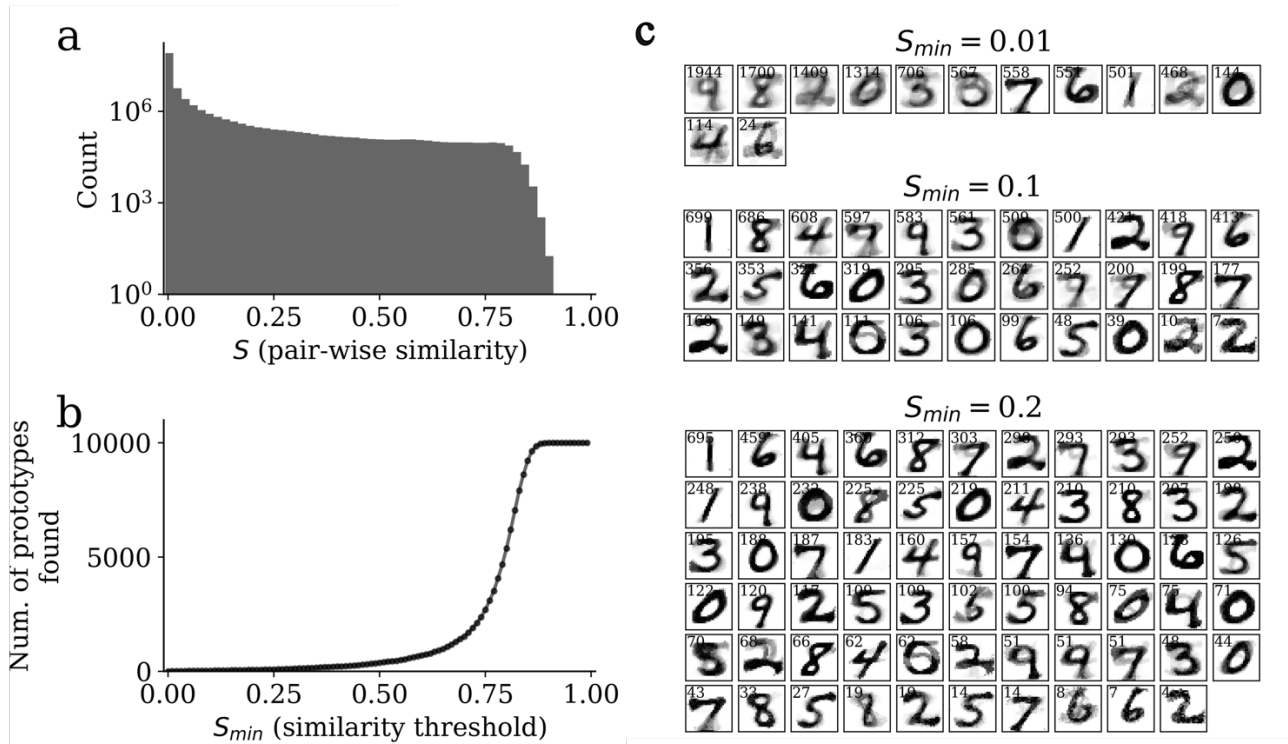

*Figure 11. Prototype extraction. (a) Distribution of pair-wise similarities (S) of HID attractor representations (T = 300 ms) with a clear mode close to zero and only a smaller fraction of high values. (b) The relationship between similarity threshold, $S_{min}$, and the number of prototypes found by grouping attractor representations into unique prototypes. (c) The prototypes found for $S_{min} = 0.01$ (upper), 0.1 (middle), and 0.2 (right) by averaging the input reconstructions from the z-traces of the INPRC population.*

**5.5 Associative memory improves the robustness of representations**

The aim of the following experiments was to compare the capacity of our models on three associative memory tasks – (1) pattern completion, (2) perceptual rivalry, and (3) distortion resistance. Considering that a significant fraction of the image is corrupted, each task is considerably harder than the (clean) MNIST test dataset. We tested if the full network with recurrent projections can produce robust representations by removing the corruptions introduced. Given the highly sparse irregular firing dynamics of the *SpspkFull* model, we sought to examine if the network can handle associative memory tasks and perform competitively with the *SpkFull* and *RateFull* model. We qualitatively assessed the evolution of *INP* and *INPRC* population outputs of the *SpspkFull* model after stimulating with one example pattern from each task (difficulty level = 0.6). For this, we visualized (Fig. 12) the raw spiking activities ($s_j$) and the short-term filtered z-traces for the *INP* and *INPRC* populations from the network at regular intervals of 20 ms after the pattern onset. For the *INP* population, we used the $z_i$-traces of feedforward projection and for the *INPRC* population, we used the $z_j$-traces of feedback projection.

For the pattern completion task (Fig. 12a), the input image was covered with a gray bar on the top covering around 8 pixels, which can be seen in the filtered *INP z*-traces (2$^{nd}$ row in Fig. 12a). From the feedforward-driven representations coming from the partial input, the associative memory model needs to recover the original memory pattern. In the initial period with the feedforward-driven representations (0-100 ms), the reconstructed images show corrupted content with traces of the top bar (*INPRC z*-traces; 60-120 ms). However, in the later phase driven exclusively through recurrent associative memory (150-300 ms), the reconstructed images reflect the convergence to a cleaned version of the corresponding digit completing the pattern (*INPRC z*-traces; 150-300 ms). The attractor-driven image reconstructions appear closer to the prototypical digit compared to the feedforward-driven image reconstructions.

The visualizations obtained in the perceptual rivalry (Fig. 12b) and distortion resistance (Fig. 12c) tasks showed similar results. The perceptual rivalry task involved presenting the network with an image combined in a smaller fraction with another rival image, as can be seen from the filtered *INP z*-traces (2$^{nd}$ row in Fig. 12b). The associative memory needs to converge to the original image representation (pattern with the strongest activation) and "win-over" the rival image. The feedforward-driven reconstructions (*INPRC z*-traces; 60-120 ms in Fig. 12b) show faithful reconstructions of the image and the rival image. However, the attractor-driven reconstructions (*INPRC z*-traces; 150-300 ms in Fig. 12b) show that the original digit is completely recovered, i.e. the convergence to the prototypical digit without traces of the rival image.

The distortion resistance task involved presenting the network with images corrupted with various distortions and the associative memory network needs to remove the distortions and recover the original pattern. The feedforward-driven reconstructions (*INPRC z*-traces; 60-120 ms in Fig. 12c) illustrate a highly distorted reconstruction image corrupted by the input noise. However, the attractor-driven reconstructions (*INPRC z*-traces; 150-300 ms in Fig. 12c) reflect the convergence to the prototypical digit without any noise from the input.

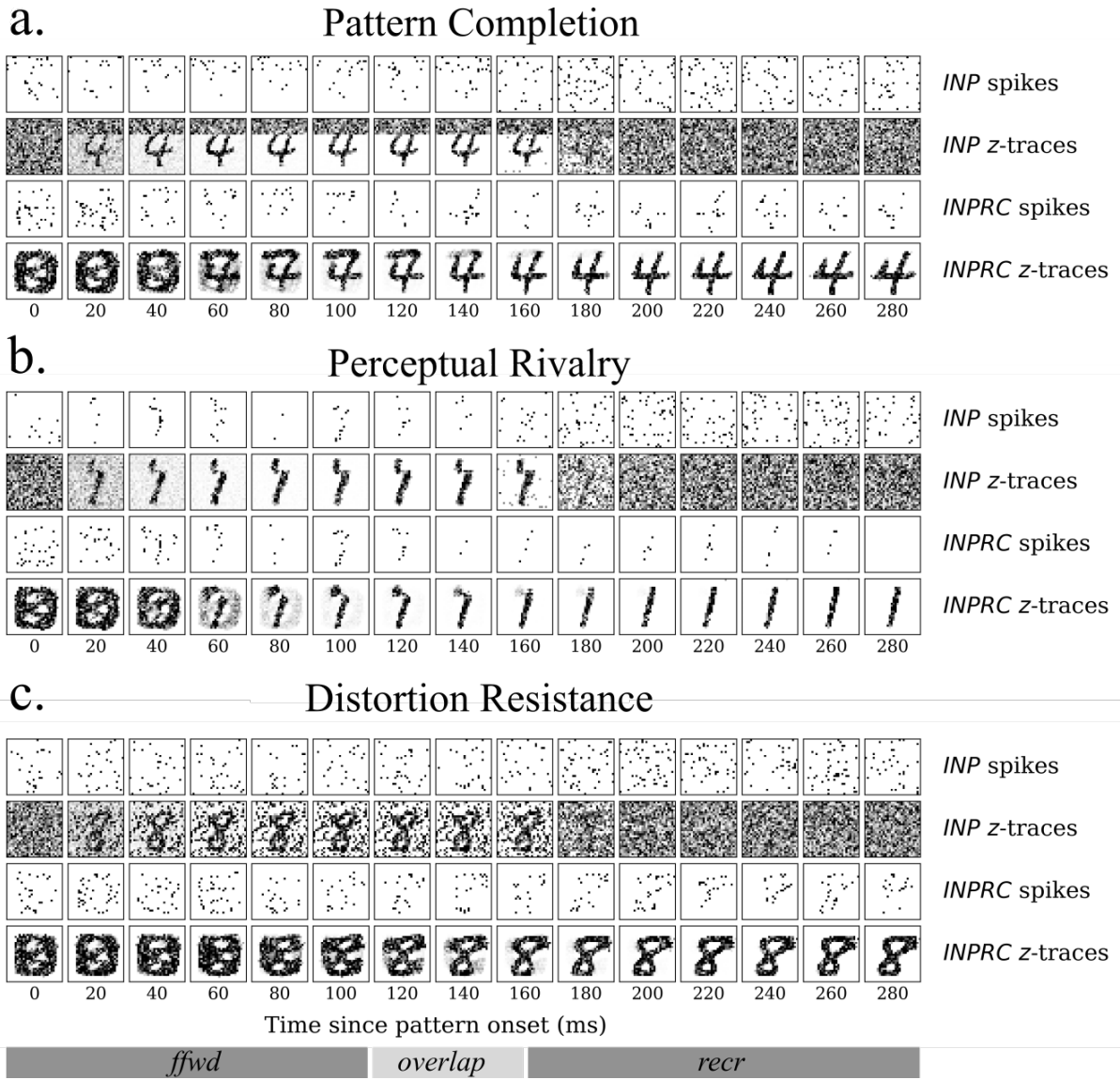

*Figure 12. Time course of attractor representations in the (a) pattern completion, (b) perceptual rivalry, and (c) distortion resistance tasks. The spikes and z-traces of INP and INPRC populations for one example pattern are shown for each task (T=300 ms; similar to the setup in Fig. 6). The INP population is driven by the spiking inputs from the corrupted image (T = 0-150 ms) with (a) top gray bar, (b) occluded partially by another rival image, and (c) randomly occurring black noise. The INPRC population shows the reconstructed image in the feedforward-driven phase (INPRC z-traces; T = 60-120 ms) is the similar to the corrupted image with traces of the top bar. In the recurrent-driven phase (T = 180-300 ms) the reconstructed image is a cleaned version of the pattern and settles on a prototypical digit representation stored in the associative memory.*

Next, we sought to quantify the performance of the network using the linear classification performance. For this, we tested the six models (*RateFf, RateFull, SpkFf, SpkFull, SpspkFf,* and *SpspkFull*) on the three associative memory tasks (pattern completion, perceptual rivalry, and distortion resistance), each on 5 difficulty levels ($N = 1000$ samples per difficulty level). The performance comparison for the models ($n = 5$ runs) is shown in Fig. 13. We observed two main results from the experiment: (1) The sparsely spiking models (*SpspkFf, SpspkFull*) perform very closely to their corresponding rate (*RateFf, RateFull*) and spiking (*SpkFf, SpkFull*) models on all the associative memory tasks and on all difficulty levels. Given the highly sparse irregular spiking activity of the models, the model performance robustly recapitulates the functionality of the rate-based model. (2) The full network models (*RateFull, SpkFull, and SpspkFull*) demonstrate a tendency for performance improvement at the high levels of task difficulty (>0.4) when compared to the feedforward-only models (*RateFf, SpkFf,* and *SpspkFf*) models. The associative memory function of the recurrent projections provides robustness to feedforward-driven representations, and this becomes clearly more beneficial in difficult settings when challenging out-of-training-set image samples are presented. This effect was quite pronounced in the pattern completion task and moderately so in the perceptual rivalry and distortion resistance tasks.

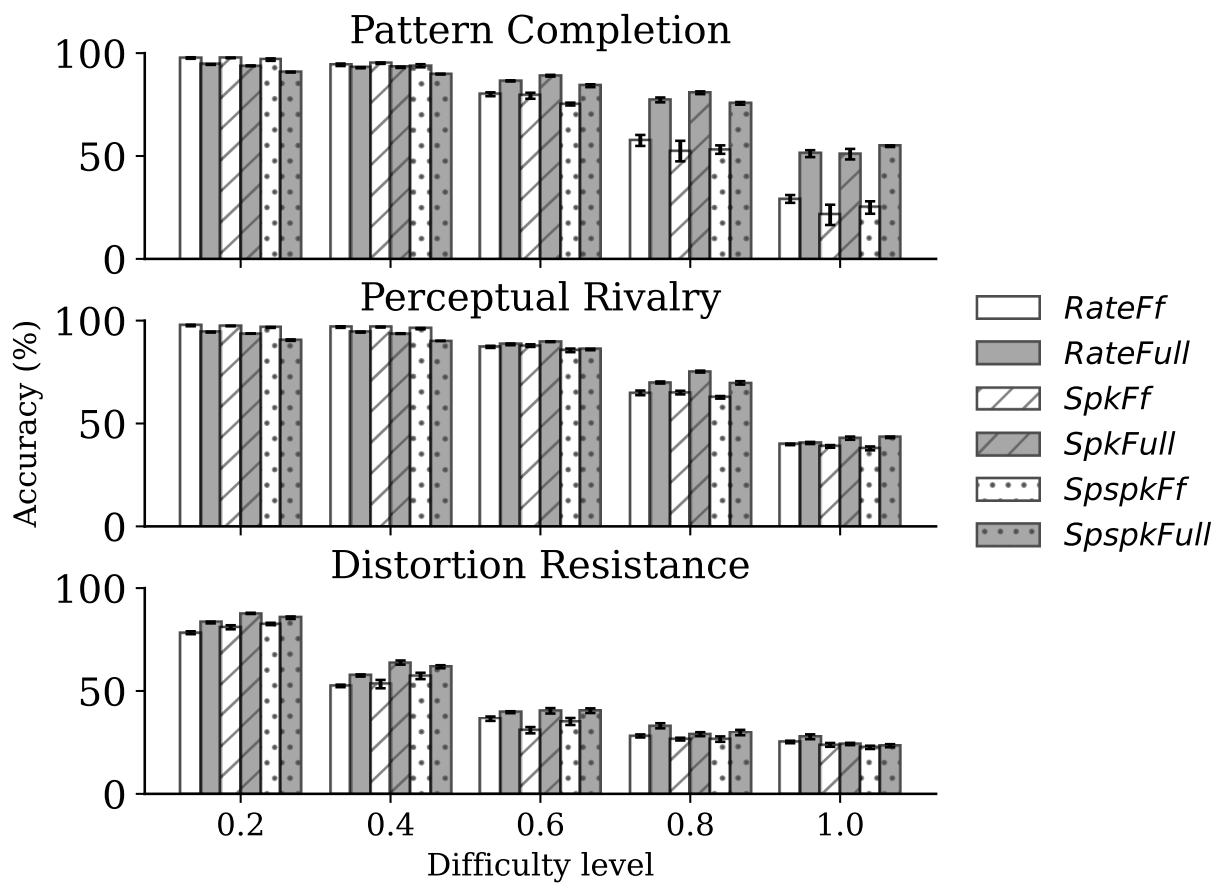

*Figure 13. Comparison of classification performance on associative memory tasks. The sparsely spiking models (SpspkFf, SpspkFull) perform very closely to the rate (RateFf, RateFull) and spiking (SpkFf, SpkFull) models in all cases. For low difficulty levels (<0.4), the full network models (RateFull, SpkFull, and SpspkFull) do not offer a clear advantage (sometimes performance slightly worse) compared to their corresponding feedforward-only models (RateFf, SpkFf, and SpspkFf) models. However, there is a clear trend of improvement for the full models compared to the feedforward-only*

*models once the difficulty level is above 0.4 in all associative memory tasks. The error bars are standard deviation from n=5 runs.*

## 6 Discussion

We introduced a novel multi-population SNN model with cortex-like modular architecture based on a stochastic Poissonian spike generation process that acts in synergy with brain-like learning and structural rewiring. We systematically evaluated and compared different variants of the basic model and showed that the sparsely spiking full (*SpspkFull*) model recapitulates many of the functionalities of the non-spiking rate-based models and demonstrated the advantages of recurrent associative memory models over feedforward-only models. Crucially, all six models were simulated within the same BCPNN implementation by modifying the parameters listed in Table 2. Hence, moving from rate-based to spiking and sparsely spiking networks needs only minor changes in the network parameters, suggesting a continuum from abstract non-spiking to more detailed spiking variants for the brain-like modeling framework presented here. Our discrete-time analog of Poisson spike generation mechanism is arguably simpler than leaky-integrate-and-fire (LIF) models, but it still recapitulates the *in vivo* irregular cortical pyramidal spiking patterns with realistic firing rates. We position our spiking neuron model as an intermediary, bridging the gap between artificial neural networks with simplistic neuron-like units (e.g., rectified linear units or sigmoidal activation functions) and biologically motivated spiking neuron models (e.g., LIF, Hodgkin-Huxley neurons). Building an analogous network with LIF neurons is a logical next step and there are strong indications that the model should perform similarly. For example, in some of our previous work focused on modeling memory function using an analogous modular recurrent network with columnar architecture and excitatory-inhibitory neuron populations we demonstrated that the spiking statistics follow Poisson distribution (Lundqvist et al., 2010). Furthermore, the Hebbian-Bayesian plasticity rule employed in our model is identical to that used in earlier SNN memory models with LIF neurons (Fiebig and Lansner, 2017; Chrysanthidis et al., 2022). This consistency is achieved because the z-traces convert spikes, whether originating from Poisson neurons or LIF neurons, into temporally averaged traces, which are subsequently used for p-traces, weights, and biases.

Furthermore, the activities in the *SpspkFull* model are highly sparse, with only around 10 spikes generated at any point of time from the *HID* population with 10000 minicolumn units. We posit that the Hebbian nature of our synaptic and structural plasticity algorithms can tolerate such highly irregular sparse spiking activations, which is in stark contrast to backprop-based learning algorithms that may not be best suited to accommodate spiking neurons.

Given that the spike generation process is a stochastic sampling of the underlying probabilistic activation, we do not expect any special advantage from using spiking signals and we did not observe any such improvements in performance from our experiments. This is contrast to many SNN studies where individual spiking timing and inter-spike intervals are proposed to provide additional information content in the neural coding signal (Wunderlich and Pehle, 2021; Eshraghian et al., 2023). However, it is still unclear if neocortical neurons communicate by means of such precisely timed spiking signals (Softky and Koch, 1993; Shadlen and Newsome, 1998). One disadvantage of our approach is, however, that the sparsely spiking neurons require long running times per pattern stimulation (5-50x) to reproduce the performance of their rate-based counterparts. This could possibly be mitigated by scaling up the network so that the number of incoming synapses per neuron approximates that of mammalian cortical pyramidal neurons (around thousand to tens of thousand) (DeFelipe and Fariñas, 1992). Integrating over such large number of stochastic sparsely spiking pre-synaptic inputs would provide a more robust summed synaptic input signal and lead to faster convergence during learning. We could also expect the response time of the neurons after pattern onset to be made shorter with such biologically realistic network scale, in agreement with fast response latency of first spikes observed *in vivo* in cortical visual hierarchy (Thorpe et al., 1996). The large time constants for $z$-traces (20-50 ms), which we showed to be necessary for networks with low firing rate (Section 5.3), could also be relaxed to shorter time constants with such large-scale networks.

We expect our network to be extendable to more commonly used spiking neuron models (such as LIF) without compromising performance. Previous modeling studies showed that local excitatory-inhibitory circuits with LIF neurons produces Poissonian statistics and reproduce well many of the *in vivo* cortical neuron spiking dynamics including oscillations and synchronization effects (Van Vreeswijk and Sompolinsky, 1996; Brunel, 2000; Lundqvist et al., 2010; Rullán Buxó and Pillow, 2020). These results strongly suggest that our spike generation process can be reproduced in more biologically detailed neuron models and integrated with other biophysical mechanisms such as spike-frequency adaptation, synaptic facilitation/depression, realistic post-synaptic potentials, axonal and dendritic delays, etc. The network can also be extended into more complex architectures, most significantly, into multilayer ones with, for instance, hierarchical feature extraction. Also, inclusion of cortical laminar organization with L4, L2/3, and L5/6 layers can allow for continuous integration of feedforward, recurrent and feedback connections reminicent of the corresponding cortical functional architecture. We used different operational phases (Section 4.4) to switch between feedforward and recurrent projections in our network to avoid one projection dominating the other. Presumably, this could be solved in a biologically plausible manner with separate laminar layers with distinct neural populations that are tightly coupled within each minicolumn, as modeled by the cortical microcircuit architecture (Douglas and Martin, 2004). Another biological mechanism could be neuromodulation as a global gating signal for synapses corresponding to specific projections.

More extensive comparison with other SNNs (trained with surrogate gradients or EventProp, for instance) and brain-inspired models will be needed to test the capacity of our model against other machine learning models. Also, traditional feedforward-driven deep learning models have been showed to severely deteriorate in performance when tested on untrained noise distortions and diverge from human behavioural performance (George et al., 2017; Tang et al., 2018; Bowers et al., 2022). Further comparisons of our model with deep learning models (such as convolutional neural networks) on noise robustness and associative memory tasks can elucidate the difference between the models.

Our work represents a step towards an integration of biologically plausible spiking models with complex brain architectures and offers exciting opportunities for scalable brain-like algorithms and multi-network models. We believe this offers high potential for the next-generation neuromorphic algorithms and hardware systems.

# 7   Conflict of Interest

The authors declare that the research was conducted in the absence of any commercial or financial relationships that could be construed as a potential conflict of interest.

## 8    Author Contributions

NB: Conceptualization, Data curation, Investigation, Methodology, Software, Validation, Visualization, Writing – original draft, Writing – review & editing; AL: Conceptualization, Data curation, Methodology, Software, Supervision, Writing – review & editing; PH: Conceptualization, Methodology, Funding acquisition, Project administration, Resources, Supervision, Writing – review & editing.

## 9    Funding

Funding for the work is received from the Swedish e-Science Research Centre (SeRC), Digital Futures, European Commission H2020 program, Swedish Research Council 2018-05360, Vetenskapsrådet (VR2016-05871 and VR2018-05360), the European Commission Directorate-General for Communications Networks, Content and Technology grant no. 101135809 (EXTRA-BRAIN)

## 10    Acknowledgments

The computations were enabled by resources provided by the Swedish National Infrastructure for Computing (SNIC) at the PDC Center for High Performance Computing, KTH Royal Institute of Technology, partially funded by the Swedish Research Council through grant agreement no. 2018-05973. We would like to thank Swedish e-Science Research Centre (SeRC), Digital Futures, Vetenskapsrådet, EXTRA-BRAIN for generous funding of the project.

## 11    Data Availability Statement

Publicly available datasets were analyzed in this study. The MNIST data can be found here: https://yann.lecun.com/exdb/mnist/. The code used to run the experiments is available publicly on GitHub and can be accessed here: https://github.com/nbrav/BCPNNSim-Frontiers2024.